\documentclass{article}

\usepackage{graphicx}
\usepackage{caption}
\usepackage{subcaption}
\usepackage[normalem]{ulem}

\usepackage{hyperref}

\usepackage[accepted]{icml2024}

\usepackage{amsmath,amsfonts,bm}

\def\eqref#1{equation~\ref{#1}}

\def\1{\bm{1}}

\def\rr{{\textnormal{r}}}

\def\vc{{\bm{c}}}

\def\vv{{\bm{v}}}
\def\vw{{\bm{w}}}

\def\mM{{\bm{M}}}

\DeclareMathAlphabet{\mathsfit}{\encodingdefault}{\sfdefault}{m}{sl}
\SetMathAlphabet{\mathsfit}{bold}{\encodingdefault}{\sfdefault}{bx}{n}

\newcommand{\R}{\mathbb{R}}

\newcommand{\softmax}{\mathrm{softmax}}

\DeclareMathOperator*{\argmin}{arg\,min}

\usepackage{amsmath}
\usepackage{amssymb}
\usepackage{mathtools}
\usepackage{amsthm}

\usepackage[capitalize,noabbrev]{cleveref}

\newcommand\abs[1]{\vert#1\vert}

\theoremstyle{plain}

\theoremstyle{definition}

\theoremstyle{remark}

\usepackage[textsize=tiny]{todonotes}

\icmltitlerunning{DNCs Require More Planning Steps}

\begin{document}

\twocolumn[
\icmltitle{DNCs Require More Planning Steps}




\begin{icmlauthorlist}
\icmlauthor{Yara Shamshoum}{yyy}
\icmlauthor{Nitzan Hodos}{yyy}
\icmlauthor{Yuval Sieradzki}{yyy}
\icmlauthor{Assaf Schuster}{yyy}

\end{icmlauthorlist}

\icmlaffiliation{yyy}{Department of Computer Science, Technion- Israel Institute of Technology, Haifa, Israel}

\icmlcorrespondingauthor{Yara Shamshoum}{yara-sh@campus.technion.ac.il}
\icmlcorrespondingauthor{Nitzan Hodos}{hodosnitzan@campus.technion.ac.il}
\icmlcorrespondingauthor{Yuval Sieradzki}{syuvsier@campus.technion.ac.il}

\icmlkeywords{Machine Learning, ICML}

\vskip 0.3in
]



\printAffiliationsAndNotice{}  

\begin{abstract}
Many recent works use machine learning models to solve various complex algorithmic problems. 
However, these models attempt to reach a solution without considering the problem's required computational complexity, which can be detrimental to their ability to solve it correctly.
In this work we investigate the effect of computational time and memory on generalization of implicit algorithmic solvers. To do so, we focus on the Differentiable Neural Computer (DNC), a general problem solver that also lets us reason directly about its usage of time and memory.
In this work, we argue that the number of planning steps the model is allowed to take, which we call ”planning budget”, is a constraint that can cause the model to generalize poorly and hurt its ability to fully utilize its external memory.
We evaluate our method on Graph Shortest Path, Convex Hull, Graph MinCut and Associative Recall, and show how the planning budget can drastically change the behavior of the learned algorithm, in terms of learned time complexity, 
training time, stability and generalization to inputs larger than those seen during training.


\end{abstract}
\section{Introduction}
\label{sec-intro}



Over the past few years, Deep Neural Networks (DNNs) have made significant advancements in various fields, including computer vision, audio analysis and speech recognition, as well as generating art and text with human-like accuracy. However, a major challenge still persists: the ability to generalize to unseen inputs. When a DNN is trained on a specific training set, its performance often decreases when presented with inputs that are outside the distribution of the training set. This can be attributed to various factors such as sparse input distributions, outliers, and edge cases. To address this issue and improve generalization, DNNs are now being trained on increasingly large datasets. For instance, in the field of Natural Language Processing (NLP), dataset sizes can reach billions and trillions of tokens \cite{kudugunta2023madlad400,wang2023diffusiondb}.

A potential solution to the issue of generalization lies in algorithms. Algorithms are designed to solve a problem for all possible cases rather than simply approximating a function. Instead of learning to approximate a function, we learn to generate an algorithm which consists of a series of steps that modify the internal state, memory, or external interfaces to ultimately achieve the desired result. The underlying assumption is that if a high-quality algorithm is discovered, it will inherently generalize to all cases. We refer to this concept as Algorithmic Reasoning in this paper.

There are multiple examples of algorithmic reasoning, which can be implemented in an explicit or an implicit manner.
In the explicit approach, the model's task is to output a description of the algorithm it has learned. Examples include AlphaTensor \cite{AlphaTensor}, in which the model learns to find general matrix multiplication algorithms for various matrix sizes; code generation models such as \cite{alpha_code}, and Large Language Model (LLM) that are able to generate a piece of code solving a task described in free text \cite{shinn2023reflexion}.

In the implicit approach, the processor learns to output actions that work for a specific input instance of the problem. To run the algorithm, we must run the model. This way, the model learns to perform the algorithm rather than describe it; the model's weights, internal representation space, and architecture comprise the learned algorithm. Examples include \cite{interfaces_paper, veličković2020pointer, neuralrandomaccess, graves2014neural} 

An important example of this approach is the Differentiable Neural Computer model \cite{Graves2016}, which is the focus of this work.
In brief, the DNC is a Recurrent Neural Network (RNN) based on a differentiable implementation of a Turing Machine, extending previous work \cite{graves2014neural}.
Featuring an LSTM with a memory matrix, the DNC can model algorithms that interact with external memory, handling tasks like copying, sorting, and graph problems.

The DNC processes inputs by iteratively absorbing a sequence of vectors, storing them in memory, and executing memory operations for task-specific outputs. It has several addressing mechanisms which allow it to allocate new cells, update existing ones or lookup a specific cell by its content. Its operation spans three phases: input, planning, and answering. Initially, it receives input, then undergoes $p$ planning steps for processing—a number previously limited to zero or just 10 in more complex tasks—and finally produces the output in the answering phase.

The implicit approach is often used to directly solve a problem without considering its time and space complexity. In general, analyzing the time complexity of an algorithm learned by a neural network can be difficult, as many factors contribute to the complexity: the optimization process, the internal representaion space etc. To make the analysis easier, we focus on DNC, whose structure lets us directly reason about time complexity and memory utilization.

\subsection{Our Contributions}

In this work, we bring a fresh perspective to DNCs, and algorithmic solvers in general, by exploring them from a perspective of computational complexity. We demonstrate the crucial role of choosing a correct planning budget on the model's ability to generalize well across various algorithmic tasks. Our findings underline the limitations of the standard $p(n)=10$ planning budget, and strongly demonstrate how simply choosing a correct planning budget can drastically improve performance. We provide strong empirical evidence for the impact of the planning budget on the behavior of learned algorithms on multiple algorithmic problems, including Shortest Path, Mincut, Convex Hull and Associative Recall. Additionally, we address the long-standing challenge of performance drop when extending DNCs external memory to support larger inputs, by identifying cause of the problem and then proposing a novel technique to overcome it. Finally, to address training instability, we propose a novel method that incorporates a stochastic planning budget, encouraging the learning of more abstract algorithms that generalize effectively. Our results extend beyond DNCs, as they describe basic principles of how time and memory resources should be utilized and applied for Algorithmic Reasoning in general.

The paper is structured as follows: Section \ref{related-work} overviews related work; Section \ref{sec-method} details our method and its complexity theory motivation; Section \ref{sec-exp} presents our experiments; and Section \ref{sec-conclusion} concludes the paper.

\subsection{Related Work}
\label{related-work}

\paragraph{Memory Augmented Neural Networks} 
Memory-augmented neural networks (MANNs) are a class of neural network architectures that incorporate an external memory structure enabling it to store and access important information over long periods of time. 
The Differentiable Neural Computer (DNC) is one such network that has been shown to be good at a variety of problems \cite{Graves2016, rae2016scaling}. Since the DNC's introduction, many researchers have tried to improve this design. \cite{franke2018robust} improved it specifically for question answering, while others have suggested changes to improve its overall performance. \cite{csordás2022improving} pointed out some issues with the DNC design and proposed fixes. \cite{augmenting_dnc_2021} suggested separating the memory into key-value pairs. In another work \cite{statespace2021}, the authors tried to encourage loop structures in the learned algorithm by constraining the state-space of the controller. There is also evidence that making the network more sparse can help with generalization and efficiency on bigger tasks \cite{rae2016scaling}. Others propose new computational architectures such as the Neural Harvard Machine \cite{nhm}.
None of these works specifically targets the impact of the planning phase on the performance of DNC.

\paragraph{Adaptive Computation Time} 
Adaptive computation time is an important aspect in solving algorithmic tasks, as more complex instances naturally require more time to solve. 
Adaptive Computation Time (\cite{graves2017adaptive}) are RNNs that incorporates a neural unit to allow the model to dynamically change the number of computational steps.
\cite{bolukbasi2017earlyexit} present Adaptive Early Exit Networks which allow the model to exit prematurely without going through the whole structure of layers.
In the context of memory-augmented neural networks, similar ideas have been proposed. \cite{reasonet}  introduces an iterative reasoning mechanism, enabling the model to refine its predictions. \cite{banino2020memo} utilizes the distance between attention weights attending the memory as a measure of how many more memory accesses the model needs. They do so by incorporating an additional unit that outputs a halting probability, which is trained using reinforcement learning.
These works, though very relevant to the claims in our paper, do not prove that adaptive computation times are a requirement.
In our work, we directly deal with the large impact the duration of computation has on the model's performance.
Allowing the model to choose its own computation time fits well with our claims in this paper, though we show that even a naively chosen planning budget already improves the model's performance substantially, without the need to alter the training procedure or add new neural modules to the model.

\begin{figure*}[!htbp]
    \centering
    \includegraphics[width=0.85\linewidth]{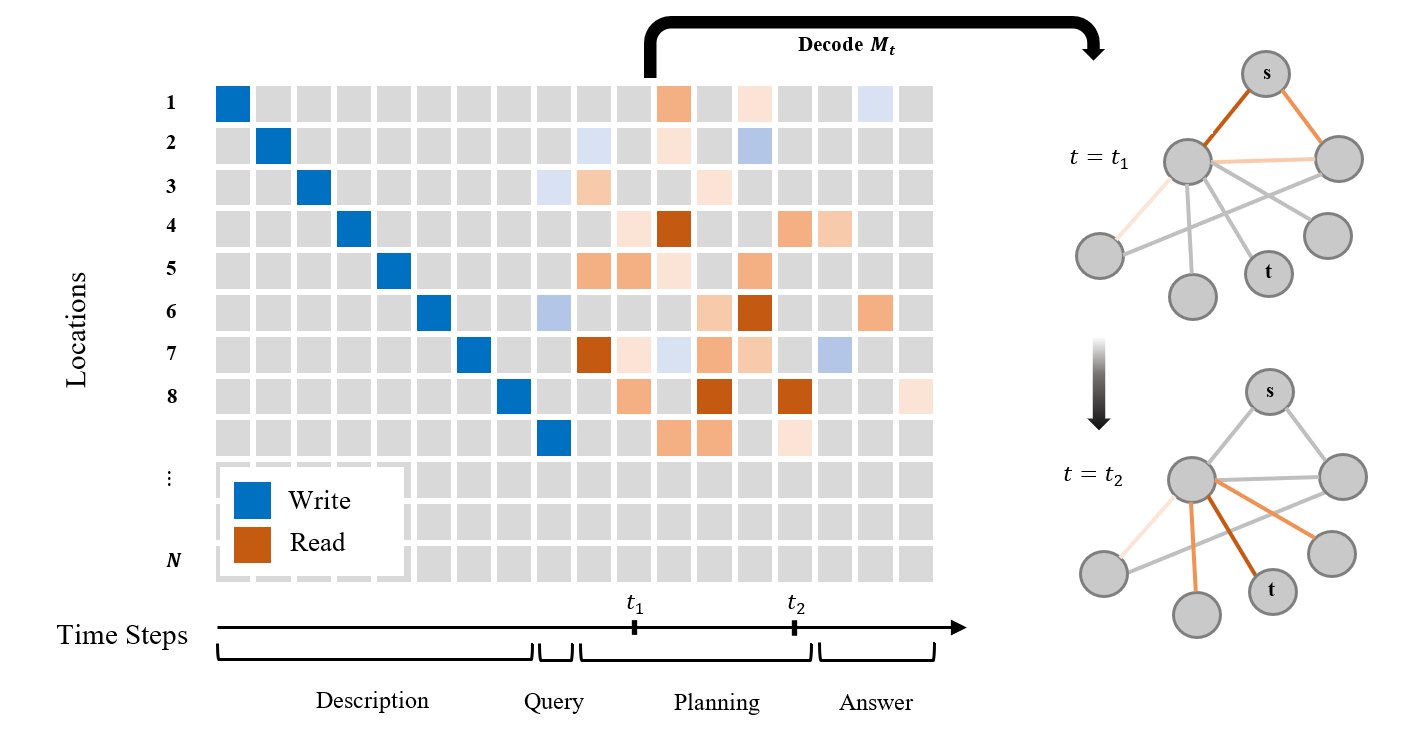}
    \caption{\textbf{An Example of a DNC Forward Pass on an Input of the Shortest Path Task}. The DNC maintains read (\textcolor{orange}{\textbf{orange}}) and write (\textcolor{blue}{\textbf{blue}}) distributions over a memory with $N$ cells. The process begins with the \textbf{description phase}, where the model receives the input, in this case graph edges, and writes them to memory. Then in the \textbf{query phase} the model is given the source and target nodes $(s,t)$, written to memory as well. Next, during the \textbf{planning phase}, the model does not receive any new external input, but can access and update its memory. Finally, in the \textbf{answer phase}, the model outputs the edges that form the calculated shortest path. 
    Decoding the read distribution during the planning phase, can provide insight to how the model traverses the graph in order to find the shortest path. By using the write distribution from the description phase, we can infer where each edge is saved in memory. This allows us to plot the read distribution over these locations during the planning phase on the graph itself, visualizing how the model locates its target. 
    }
    \label{fig:dnc}
\end{figure*}
\section{DNC Recap}
\label{sec-dnc}

The Differentiable Neural Computer (DNC) \cite{Graves2016} is a memory-augmented neural network, based on a recurrent neural representation of a Turing Machine. A DNC consists of a controller coupled with an external memory $\mM\in\R^{N\times C}$, where $N$ is the number of memory cells and $C$ is the size of a memory cell. The external memory can be accessed by the controller through different addressing mechanism, allowing the model to write to unused memory cells, update cells, and lookup specific cells based on their content. The external memory allows the DNC to be a general problem solver, and it was shown to work successfully on a wide range of tasks such as sorting, question answering and more.

The controller learns to interact with the memory using $m$ fully differentiable read heads and a single write head, allowing the model to learn to utilize $M$ through an end-to-end training process.
Each read head $R^i$ accesses the memory at every timestep $t$ by generating weights over the address space $\vw^{r,i}_t \in \R^N$, with the read value being $\rr^i_t=\mM_t ^\top\vw ^{r,i}_t$. Similarly the write head $W$ updates the memory through a generated write distribution $\vw^w_t \in \R^N$.

The controller of DNC acts as a state machine, managing memory access through generated signals.
Its input is a concatenation of the input vector at time step $t$ denoted $x_t$, as well as the $m$ values read from the memory at the previous timestep, $\vv_{t-1}^i$.
The output of the controller is mapped into two vectors: a control vector $\xi _t$ used to control the memory operations, and an output vector $\nu _t$ used to generate the final output $o_t$. 

The process of generating an answer sequence by DNC can be divided into multiple phase.  First, during the \textbf{description phase}, the network sequentially receives a description of the problem instance. For example, in the Graph Shortest Path scenario, this description could be the set of edges in an input graph. Subsequently, in the \textbf{query phase}, an end-of-input token $\langle\text{eoi}\rangle$ is presented followed by an optional query. In Graph Shortest Path, the query is the source and target nodes $(s,t)$. The \textbf{planning phase} which allows the model to access its state and memory by providing a zero-vector for \( p \geq 0\) time steps. Finally, in the \textbf{answer phase} initiated by an answer token $\langle\text{ans}\rangle$, the model outputs the answer sequence $y$. An example of how DNC interacts with its memory during the different phases is provided in Figure \ref{fig:dnc}.

\section{Method}
\label{sec-method}
\subsection{Motivation}
\label{sec-motivation}

In complexity theory there are multiple lower bound results showing that many problems cannot be solved by a Turing Machine in constant time. For example, \cite{lower_bound_shortest_path} show a $\mathcal{O}\left(\log^3(n)\right)$ lower bound for the time complexity of Graph Shortest Path, though for a more complex computational model than that of the DNC.
The DNC, as a computational model, is equivalent to a Turing Machine. However we don't know if this equivalence gives the DNC better time complexity due to its learned embedding spaces. The DNC model could find an efficient representation of the data that allows, for example, to query a graph's node as well as its neighbors in a single read operation. In this way the model can reduce the number of read/write operations it has to take to gather the same data.
This can be achieved by encoding ``neighborhood information'' in the embedding value stored in a single memory cell, representing an efficient, aggregated piece of information about the input.

This implies that we can assume the embedding space allows the model a multiplicative ``efficiency factor'' $k$, compared with a Turing Machine. Consequently, the DNC's asymptotic time complexity is bounded from below by $T(n)/k$, where $T(n)$ is the asymptotic time complexity of some Turing Machine solving the same problem.
However, the amount of information the internal representation can learn to represent is finite; in fact we can assume its implicit dimensionality is much smaller than its embedding dimension, as is often the case for learned embedding spaces in DNNs \cite{ansuini2019intrinsic}.
This implies that $k$ is bounded and not too large, meaning the runtime of the DNC is at least a linear function of the runtime of some Turing Machine.

For a problem with a lower bound $B(n)$ on the time complexity of a Turing Machine solving that problem, i.e. $B(n) \leq T(n)$, we immediately see that the time complexity of a DNC solving the same problem will be bound from below by $B(n)/k$.
In other words, for most ``interesting'' problems, a constant runtime is simply not enough, motivating a choice of an \textbf{adaptive planning budget} - given an input $x$ with description length $\lvert x\rvert=n$, we set the planning budget to be a function of the input size, $p(n)\to\mathbb{N}$.

Alternatively, one could train with a larger constant planning budget to support larger inputs. However, scaling the planning budget can be very costly, especially during training. Increasing the planning budget too far can also cause optimization problems such as vanishing gradients and training instability, as during training gradients propagate through time. 

We show that choosing a planning budget that is correct for the problem can drastically improve generalization. If the model successfully learns a good representation of the input and an abstract algorithm to process this representation, it should generalize and perform well on larger inputs.


We also find that changing the planning budget used by a DNC model has great effect on the the behavior of the learned algorithm, in terms of learned time complexity, training time, and stability.

\subsection{Generalization with DNC}
\label{sec-motivation-mem}
When attempting to generalize to larger inputs, we are immediately met with the finite size of the DNC's external memory. 
As the memory size provided during training is limited, failing to generalize to larger inputs could be attributed either to the algorithm not supporting these input sizes, or simply the lack of memory to run the algorithm properly. To alleviate this constraint, we turn to discussing the usage of larger memory.

 Unfortunately, training with large memories is costly, as the training time of DNC scales with it \cite{rae2016scaling}. 
An alternative approach by \cite{Graves2016} uses a larger external memory during inference only, and was shown to perform well for simple tasks.
However, \cite{statespace2022} illustrated a decay in performance with an extended memory, especially for more complex tasks.

We show that this decrease in performance can be attributed to the over-smoothing of the scores of the content-based access mechanism of the DNC. To address this issue, we propose a solution involving the reweighting of these scores, by introducing a \textbf{temperature recalibration parameter}.

When dealing with computationally interesting problems, their required space complexity is often input dependent as well, as information has to be stored to memory. This implies that a constant memory size cannot be optimal for all inputs for such problems, no matter what temperature we use for recalibration. Hence, we are also motivated to use the DNC with an \textbf{adaptive memory} - given an input $x$ with description length $\lvert x\rvert=n$, we set the memory size to be a function of the input size, $m(n)\to\mathbb{N}$.

\section{Experiments}
\label{sec-exp}
\begin{figure}[!htbp]
  \centering
   \includegraphics[width=0.85\linewidth]{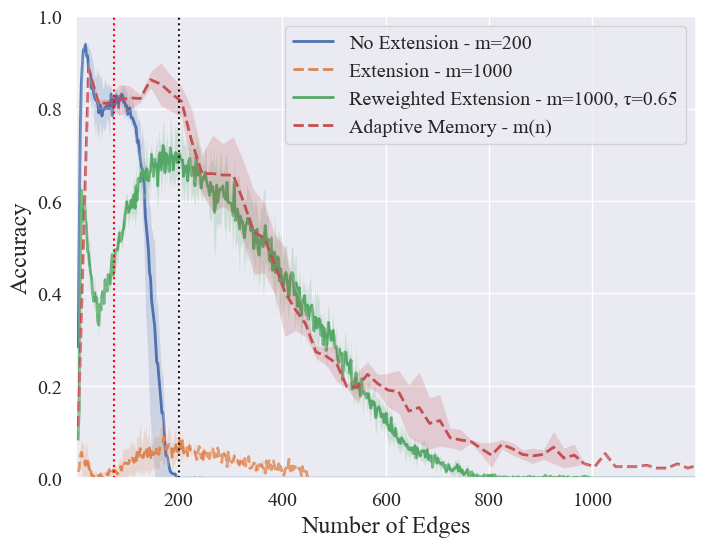}
    \caption{\textbf{Effect of Different Memory Extension Techniques on Generalization} - Evaluated on Graph Shortest Path task with $p(n)=n$. Graphs seen during training have at most 75 edges, marked in \textbf{\textcolor{red}{red}}. The memory size used for training is 200 cells, marked in \textbf{black}. A performance drop occurs around the original memory size of $m=200$ when attempting to generalize without memory extension. Extending the memory five times to $m=1000$ results in near-zero accuracy on all input sizes. Introducing our reweighting technique with $\tau = 0.65$ enables generalization to much larger inputs. Finally, using an adaptive memory during inference allows generalization while maintaining high accuracy on smaller inputs too.}
    \label{fig:mem-extension-demo}
\end{figure}
\subsection{Training}
In our analysis, we examine how DNCs trained with different planning budgets generalize to inputs larger than those seen during training. As outlined in our motivation, the complexity of a problem significantly impacts the correct choice of budget. Consequently, we evaluate our models on Shortest Path, Convex Hull and MinCut as examples of ``interesting'' problems who cannot be optimally solved with an online algorithm with constant latency.

Moreover, we evaluate the models on the Associative Recall task, which can be solved with an online algorithm with negligible latency. Associative Recall requires saving the input to memory in some form, hence ideally the optimal query time would be gained by constructing an index. As we used input values in base 10 of up to 5 digits, querying the answer from a constructed index that is a 10-ary tree would take mere 5 operations. 
We can conclude that an additional planning time over the baseline is unnecessary for optimal solution of the problem, which is why we consider it an easy problem.

We train our models with various planning budgets including the baseline of $p(n)=10$, larger constant planning budgets and adaptive planning budgets. 
The specific constants tested depend on the problem, with the maximum being the memory size to guarantee that it can fit in memory during training.
When choosing the specific function for the adaptive budgets, we rely on the known problem complexity as our guideline. We thus begin by comparing the constant budgets with a linear one $p(n)=n$, for Graph Shortest Path. Additionally, we experiment with different coefficients $p(n)=kn$ for the Convex Hull problem, testing values such as $k=0.5, 1.5$. However, we find that these variations had little effect on performance or training efficiency. For the remainder of this section, we present the results for the Convex Hull task with $k=1.5$. Further details and figures are supplied in Appendix \ref{sec-convex}.

All of the models are trained using the same constant memory size and on the same data distribution following a curriculum. We refer to Appendix \ref{sec-train-setup} for problem descriptions and training details, and Appendix \ref{sec-curriculums} for the curriculums used for training.

\begin{figure}[!htbp]
    \centering
    \begin{subfigure}{\linewidth}
        \centering
        \includegraphics[width=\linewidth]{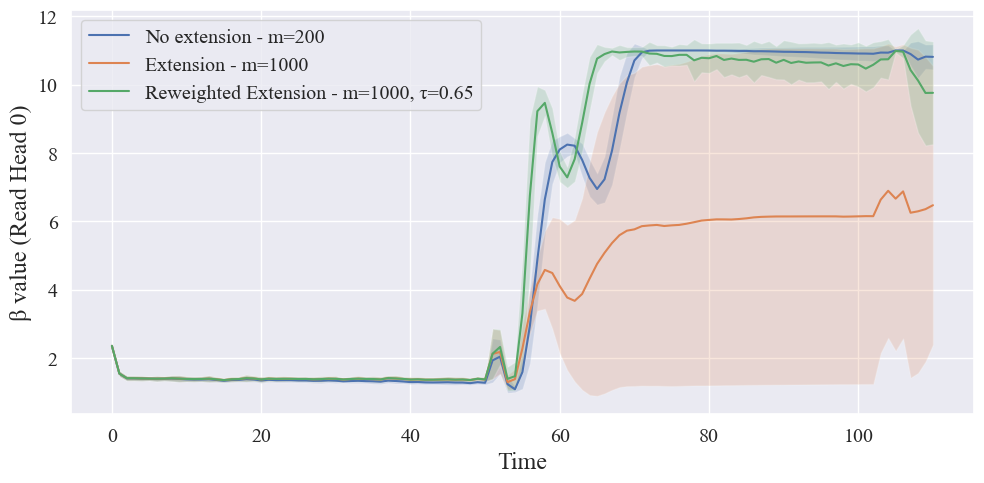}
        \caption{$\beta$ values for input size $n=50$}
        \label{fig:n=50-betas} 
    \end{subfigure}
    \hfill
    \begin{subfigure}{\linewidth}
        \centering
        \includegraphics[width=\linewidth]{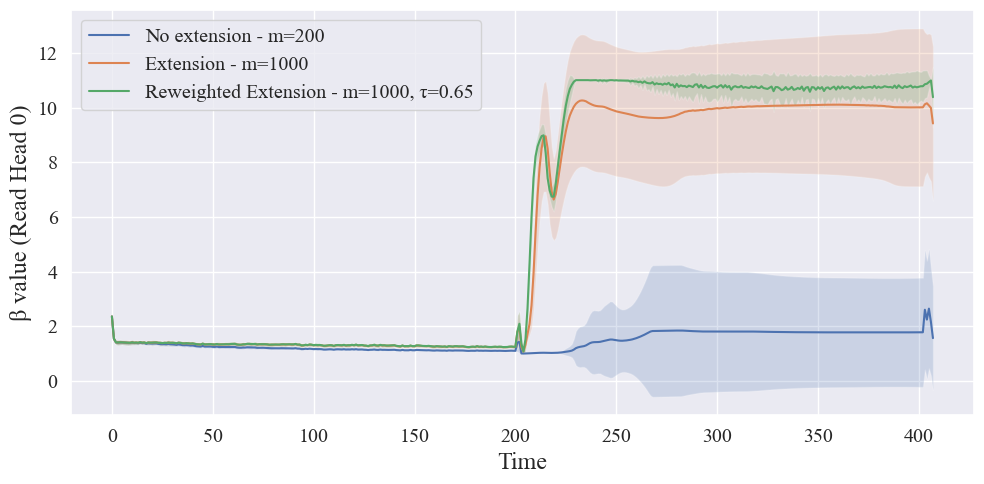}
        \caption{$\beta$ values for input size $n=200$}
        \label{fig:n=200-betas}
    \end{subfigure}
    \\[\baselineskip]
    \caption{{\boldmath\bfseries Effect of Memory Reweighting on the Strength Scalar $\beta$} - Evaluated on Graph Shortest Path with $p(n)=n$. 
    In DNC, read and write operations are smooth and as a result add noise to the memory, an effect that is more prominent when the memory is extended. As $\beta$ attempts to calibrate the smoothness of the similarity scores between the key and the memory cells, we expect that the same $\beta$ value will be optimal when using an input that is 5 times larger within a memory that is 5 times larger, as the same ratio of noise values gets into the similarity score. When applying our technique to (a) a small input and (b) a large input, the temperature reweighting recalibrates $\beta$ to be optimal for the memory used during training and the noise ratio determined by the input sizes seen during training. Hence, large inputs within the extended memory will gain performance as this ratio is matched, while for small inputs this will cause degradation in performance. We also notice how the temperature reweighting drastically reduces the standard deviation of $\beta$, which is expected as it sharpens the distribution, making it more certain.
    }
    \label{fig:betas}
\end{figure}
\subsection{Memory Extension for Generalization}
\label{sec:mem-extension}

As mentioned in Section \ref{sec-motivation-mem}, we first have to describe the degradation in performance that comes with memory extension during inference, as it prevents us from applying the DNC to inputs larger than those seen during training. 

\begin{figure*}[!htbp]
    \centering
    \begin{subfigure}{0.33\linewidth}
        \centering
        \includegraphics[width=\linewidth]{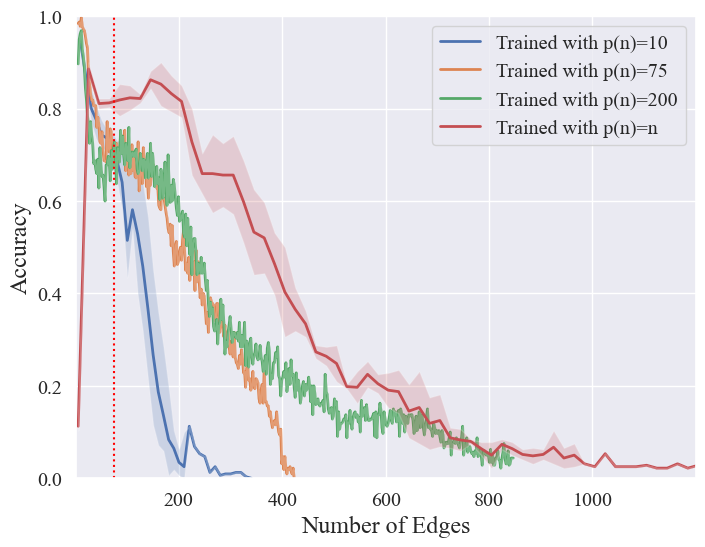}
        \caption{Shortest Path Task}
        \label{fig:sp-generalization} 
    \end{subfigure}
    \hfill
    \begin{subfigure}{0.33\linewidth}
        \centering
        \includegraphics[width=\linewidth]{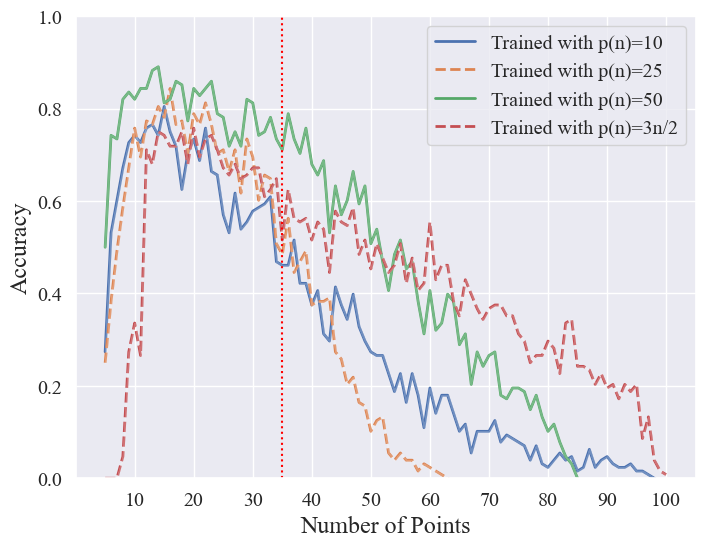}
        \caption{Convex Hull Task}
        \label{fig:ch-generalization}
    \end{subfigure}
    \hfill
    \begin{subfigure}{0.33\linewidth}
        \centering
        \includegraphics[width=\linewidth]{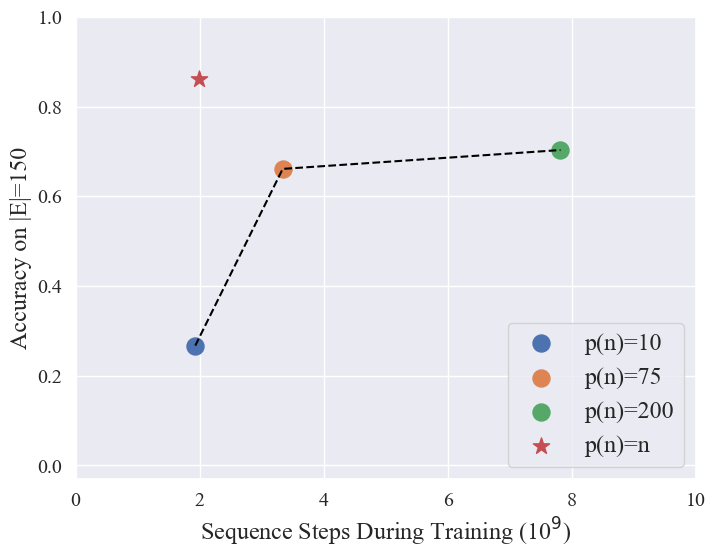}
        \caption{Training FLOPs for Shortest Path Task}
        \label{fig:flops}
    \end{subfigure}

    \caption{\textbf{Effect of Planning Budget on Generalization and Training Efficiency} - Generalization of various planning budgets on \textbf{(a)} Shortest Path Task and \textbf{(b)} Convex Hull Task demonstrates improvement with some larger budgets as well as the adaptive budget. The largest training sample is marked in \textbf{\textcolor{red}{red}}. In both tasks, the model's generalization improves over the baseline of the previously used planning budget $p(n)=10$. Subfigure \textbf{(c)} illustrates the estimated number of FLOPs for each planning budget on the Shortest Path task. The accuracy is evaluated on inputs twice the size of the largest training sample. Notably, the training of the model with the adaptive budget proves to be as efficient as the model trained with the smallest constant budget, while outperforming the model with the largest planning budget. A performance drop for very small inputs in \textbf{(a)} and \textbf{(b)} is observed, which we attribute to the training using curriculum learning. Towards the end of their training, the focus shifts to larger training samples, potentially leading to forgetting the easier ones. }
    \label{fig:generalization}
\end{figure*}
In DNC, the distribution of the content-based access is calculated in 3 steps: First, the controller produces a key vector. Then a similarity score is computer for each memory cell $\hat{\vc}$ . Finally, these scores are normalized using the softmax function $\vc = \softmax(\hat{\vc}\cdot\beta)$, where $\beta$ is a scalar strength produced by the controller at each timestep. Consequently, increasing the size of the external memory during inference produces a smoother distribution, more spread out over the larger address space than the one seen during training. We believe that this leads to the observed degradation in performance, and propose sharpening the distribution during inference to avoid this effect.

To achieve this, we recalibrate the scalar strength $\beta$ to align with the extended memory size by introducing a temperature recalibration parameter $\tau$: $\vc = \softmax(\hat{\vc}\cdot\frac\beta\tau)$. The parameter $\tau$ can be found through hyperparameter search, We found that a value of $\tau =0.85$ works well for extending the memory to double its original size, and $\tau =0.65$ allowed us to extend the memory five times.

As can be seen in Figure \ref{fig:mem-extension-demo}, this memory reweighting technique significantly mitigates the performance drop, although some accuracy degradation still occurs for smaller inputs, who had better performance when using the memory given during training. 
Since smaller inputs may use less of the extended memory, the reweighting of the mostly non-relevant memory introduces noise into the score, so this relative degradation is expected. We visualize this effect in Figure \ref{fig:betas}. 

As this degradation can be avoided by using an input dependent memory size $m(n)$, we employ an adaptive memory strategy during inference. Specifically, as the input size increases, we monitor the number of memory allocations made by the model and adaptively adjust the memory size and $\tau$ accordingly. 
For the smallest inputs we begin with a training-sized memory and a temperature $\tau=1$. Upon exceeding a threshold of $65\%$ of the available memory, we expand the memory by a factor of 2 and adjust the temperature by multiplying it with a factor of $\alpha = 0.85$. 
The optimal value for $\alpha$ can determined through a hyperparameter search. As demonstrated in Figure \ref{fig:mem-extension-demo}, this solves the potential accuracy degradation on smaller inputs.

Now, we can compare the generalization of different planning budgets without worrying that the learned algorithms lack memory when tested on large inputs.

\subsection{Planning Budget Affects Generalization}
\label{sec-generalization}

We compare different planning budgets by assessing their accuracy on inputs larger than those seen during training. Simultaneously, we assess the training efficiency of these different planning budgets by estimating the FLOPs used for training. To achieve this estimation, we track the total number of timesteps throughout the training process. Since each timestep corresponds to a single pass though the DNC model, it correlates to a fixed number of FLOPs.

As illustrated in Figure \ref{fig:sp-generalization} and \ref{fig:ch-generalization},
the model trained with the standard planning budget used in previous work, $p=10$, demonstrates poor generalization. Conversely, the model appears to benefit from a larger constant planning budget. Most importantly, the adaptive budget outperforms the baseline, and is nearly matched  only by the largest constant budget.
Concurrently, Figure \ref{fig:flops} reveals  that for the Shortest Path task,  the adaptive approach is as efficient as the baseline while outperforming all of of the constant budgets. The highest constant budget achieves the nearest accuracy at a cost of $\times 4$ training time, aligning with our computational motivation. Similarly, we refer the reader to Appnedix \ref{sec-convex} for similar results for Convex Hull, where the adaptive budget is matched only by the highest constant budget, while requiring less than half of the training time.
\begin{figure}[!htbp]
  \centering
   \includegraphics[width=0.85\linewidth]{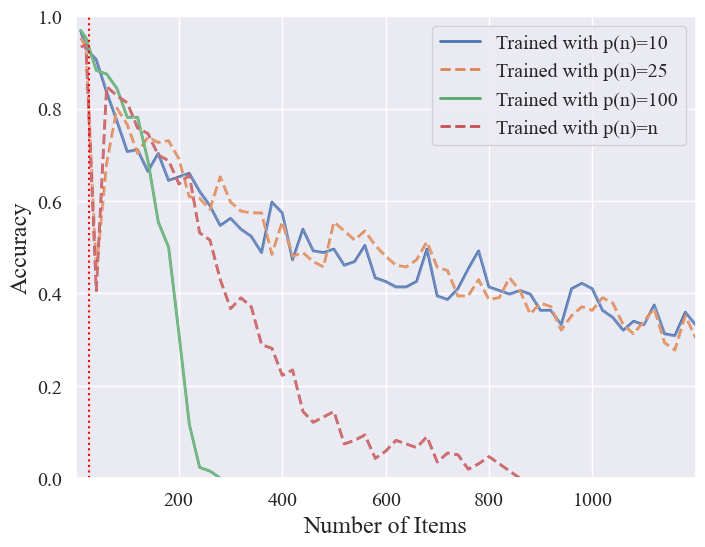}
    \caption{\textbf{Effect of Planning Budget on Generalization for Associative Recall Task} - The baseline of $p=10$ generalizes well, and the model does not benefit from the larger planning budgets. This aligns with our expectations, considering the simplicity of the task, which can be efficiently solved online. }
    \label{fig:ar-generalization}
\end{figure}
However, training with extremely long planning budgets can hurt performance. We note that the advantage of an adaptive $p(n)$ is much less prominent on MinCut, where the best-known algorithm has a time complexity of $\mathcal{O}\left(\lvert{V}\rvert\cdot\lvert{E}\rvert +\lvert{V}\rvert^2\log(\lvert{V}\rvert\right)\approx \mathcal{O}(n^2)$. We also experimented with a planning budget of size $p(n)=\lvert{V}\rvert \cdot\lvert{E}\rvert$, which proved ineffective and challenging to train. This aligns with our motivation that DNCs may struggle to learn an algorithm with an extremely large planning budget, as it involves training over very long sequences, which can introduce challenges such as vanishing gradients and training instability as DNCs are recurrent-based neural networks.
For detailed results on MinCut we refer to Appendix \ref{sec-mincut}, and for results about the effect of such quadratic budget on Shortest Path we refer to Appendix \ref{sec-quadratic}.

As for the Associative Recall task, Figure \ref{fig:ar-generalization} demonstrates the generalization of the different planning budgets. As expected, the model does not gain much by introducing an enlarged planning budget, neither constant or adaptive, as even the baseline demonstrates effective generalization for this computationally ``easy'' problem.

\subsection{Empirically Determined Planning Budget}
\label{sec-empirical-budget}
\begin{figure*}[!htbp]
    \centering
    \begin{subfigure}{0.49\linewidth}
        \centering
        \includegraphics[width=\linewidth]{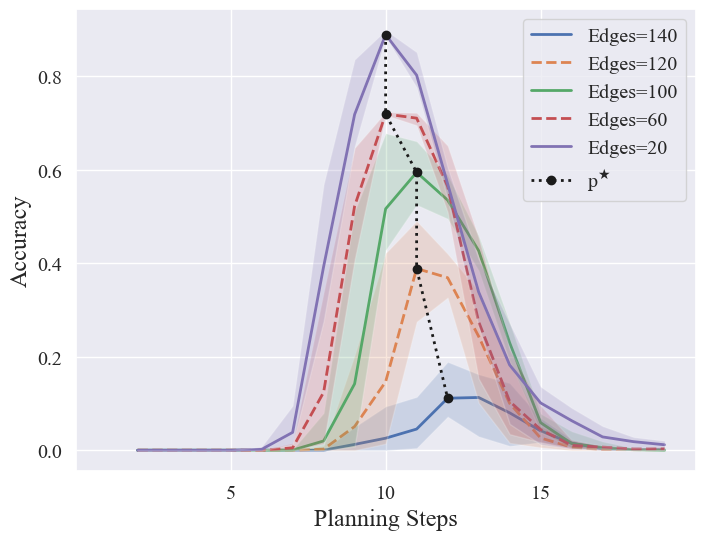}
        \caption{Model accuracy when trained with $p(n)=10$}
        \label{fig:constant-peaks} 
    \end{subfigure}
    \hfill
    \begin{subfigure}{0.49\linewidth}
        \centering
        \includegraphics[width=\linewidth]{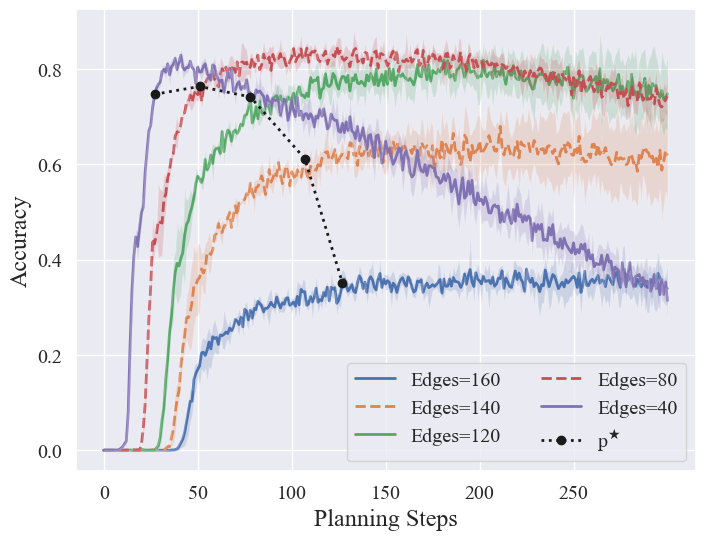}
        \caption{Model accuracy when trained with $p(n)=n$}
        \label{fig:linear-peaks}
    \end{subfigure}

    \caption{
        {\boldmath\bfseries  $A_n(p)$ for different input sizes $n$, Shortest Path} - Each colored line represents model accuracy over graphs of a chosen size, as a function of number of planning steps. Black dots denote the empirically determined planning budget $p^\star(n)$. The model trained with constant budget model only works when given $p=10\pm5$, whereas the model trained with linear budget maintains stable accuracy across various $p$ values.
        This indicates the models learned truly different algorithms.
    }
    \label{fig:peaks}
\end{figure*}
During training, the model learns an implicit algorithm whose time complexity is unknown and may differ from the planning budget used in its training. Even if this learned algorithm truly generalizes, the planning budget we use in inference might simply be too short for the learned algorithm. Instead, by granting the algorithm a larger planning budget during inference, a general algorithm could achieve better generalization, even if it was found when training with a much smaller planning budget.
Instead of choosing our planning budget in advance, we can infer the optimal budget during inference by observing the model's performance. Let $A_n(p)$ be the model's accuracy on inputs of size $n$ and a given number of planning steps $p$. For a specific value of $n$, we evaluate $A_n(p)$ for all $p\in[0,300]$. Results for Shortest Path are shown in Figure \ref{fig:peaks}, for Convex Hull in Appendix \ref{sec-convex}, for MinCut in Appendix \ref{sec-mincut}, and For Associative Recall in Appendix \ref{sec-ass}.

For the constant budget DNC, $A_n(p)$ is non-zero only near $p=10$, indicating that it will not see any benefit from a different planning budget than the one used in training. In contrast, for the adaptive budget DNC, the function $A_n(p)$ shows a phase transition. With too few planning steps, the performance is low, but after some threshold value the
performance jumps to a high level and remains there even if the number of planning steps is significantly increased. This phase transition value indicates a good choice for an empirically determined
planning budget; using more planning steps is not very beneficial, and using less is detrimental. We mark this phase transition value as $p^\star(n)$, defined as the smallest $p$ for which accuracy exceeds 90\% of its maximum value: $p^\star(n) = \argmin{\{p\mid A_n(p)>0.9\cdot \max\{A_n(p)}\}\}$.

The phase transition in $A_n(p)$ can be understood as follows: For $p \leq p^\star(n)$, the model's accuracy could be improved if given more planning, suggesting that for some inputs the learned algorithm's runtime exceeds the planning budget. For all $p \geq p^\star(n)$, accuracy plateaus, indicating that the algorithm typically concludes before reaching $p$ planning steps, with the model holding a 'finished' steady-state until the answering phase. This steady-state behavior could imply a truly generalized algorithm, as it contrasts with the less stable performance of the baseline model.

\begin{figure}[!htbp]
    \centering
        \includegraphics[width=0.85\linewidth]{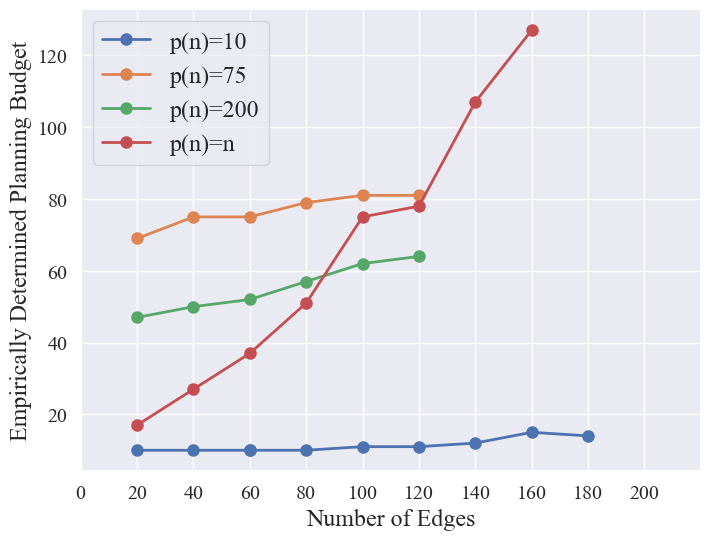}
        \caption{\textbf{Empirical Planning Budget, Shortest Path} - The empirical planning budget closely matches with the training budget $p^\star(n) \approx p(n)$, indicating the models will not benefit from additional planning steps during inference. }
    \label{fig:sp-opt}
\end{figure}

In Figure \ref{fig:sp-opt}, we observe that for the Shortest Path task, $p^\star(n) \approx p(n)$. This suggests that the models learned to execute algorithms with runtimes matching their allotted training times, and that this runtime generalizes very well into inputs much larger than those seen during training. On the contrary, more planning steps don't enhance the baseline's performance. Essentially, our results indicate the need to maintain the same planning budget during inference for optimal performance of the trained models. This emphasizes that by simply allowing the DNC an adequate planning budget such as $p(n)=n$, the DNC can learn a more "general" algorithm on more complex problems. We refer to Appendix \ref{sec-convex} for similar results for Convex Hull. Interestingly, while analyzing Figure \ref{fig:linear-peaks}, we observed some inconsistencies in the linear budget models. Despite outperforming the baseline models, they exhibited less stable accuracy across different $p$ values. For further details ans figures see Appendix \ref{sec-bad-seeds}.

\subsection{Stochastic Planning Budget During Training}
\label{sec-stability}

As we briefly mentioned in the last section, we found the training of DNC to be rather unstable, as simply changing the random seed that controls the training process could cause a model of the same planning budget to fail to generalize. Upon closer inspection, we noticed that when some seeds failed and some succeeded, only the successful ones exhibited the steady-state behavior we just presented.

We attribute this lack of steady-state behavior to an overfitting of the model to the specific planning budget used during training, as the model is unable to work with any budget different from that specific value. To prevent this, we propose adding a stochastic amount of planning steps to the planning budget during training. We call this regularization technique a "stochastic planning budget". 
Intuitively, if the model does not know when the answering phase will begin, the best strategy would be to retain the answer in memory until it finally does.
\begin{figure}[!htbp]
    \centering
    \includegraphics[width=0.85\linewidth]{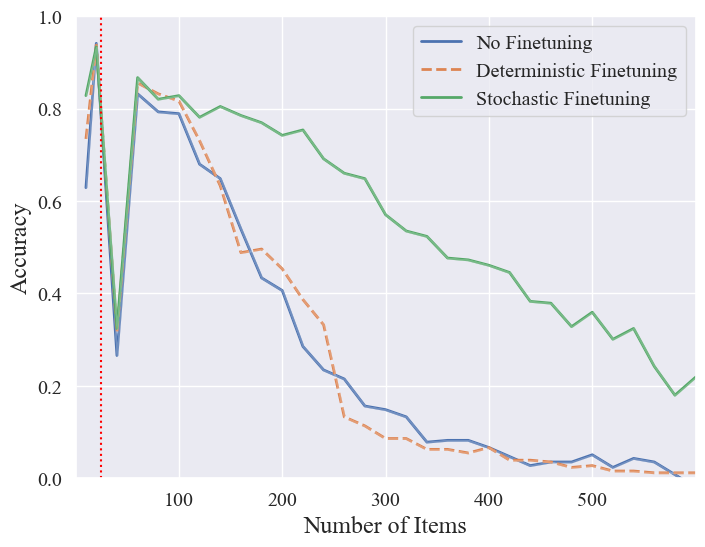}
        \caption{\textbf{Effect of Stochastic Planning Budget Fine-tuning}. Evaluated on Associative Recall, and $p(n)=n$.  the stochastic fine-tuning significantly improved the generalization of the trained model. Conversely, the deterministic fine-tuning did not exhibit any noticeable effect on generalization.}
    \label{fig:sto}
\end{figure}
To validate our hypothesis, we take initially unsuccessful models and finetune them with the same planning budget $p(n)$, but with an additional stochastic number of steps. These steps are sampled from a geometric distribution, where the probability of stopping after an additional $p(n)$ steps is at least $95\%$. 

To ensure that the gain in performance is attributed to the randomness rather the additional planning time, we compare the finetuned models to their deterministic equivalents. Namely, the models are alternatively fine-tuned with an addition of the expected value of the stochastic addition.

As can be seen in Figure \ref{fig:sto}, the stochastic planning budget allows better generalization on the same planning budget, ``fixing'' the failure we appropriated to the specific random used for this evaluation.
In addition, the comparative deterministic baseline performed poorly, proving that the gain is attributed to the stochasticity of the introduced technique.

\begin{figure}[!htbp]
    \centering
    \includegraphics[width=0.85\linewidth]{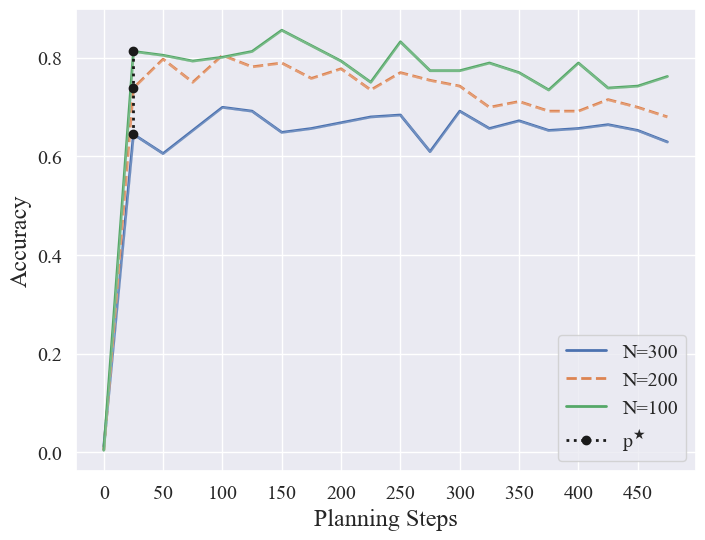}
        \caption{{\boldmath\bfseries  $A_n(p)$ for different input sizes $n$ after the stochastic finetuning} - for Associative Recall and $p(n)=n$. The stochastic fine-tuning demonstrates its efficacy in promoting the learning of retaining a steady state.}
    \label{fig:stoch-knee}
\end{figure}

In Figure \ref{fig:stoch-knee}, it is evident that after fine-tuning the model with the addition of stochastic planning steps, the model now exhibits the ability to hold a steady state. Additionally, it's worth noting that, due to time constraints, we opted to test this approach by fine-tuning for a limited number of steps. However, there is potential for further improvement with longer fine-tuning or training the model with stochastic additions throughout the entire curriculum. 

Additionally, we tested this approach on the Shortest Path task for a seed that generalizes poorly, and observed similar results. We refer to Appendix \ref{sec-bad-seeds}, for further details.


\section{Conclusion and Future Work}
\label{sec-conclusion}

In this work, we are the first of our knowledge to consider evaluating the DNC model from a computational complexity approach. While other works focus on directly maximizing accuracy, we attempt to gain general understanding of how Algorithmic Reasoners should utilize time and memory, and how they can perhaps learn to define the required resources on their own. We introduce novel techniques that are essential to evaluate DNCs, or any other memory-augmented neural network, on larger inputs. We provided strong experimental evidence that simply changing the choice of planning budget can greatly improve the performance of a DNC, and demonstrated it on multiple problems: Shortest Path, MinCut, Associative Recall and Convex Hull. We draw general conclusions regarding learned time complexity, training time and stability that may be relevant to Algorithmic Reasoners in general. For example, we show how successful algorithms tend to reach a certain steady state, holding their results in memory until they have to provide an answer. We can also hypothesize that the runtime constraint we described, which can be lifted by using adaptive planning budgets, also applies to other models for algorithmic reasoning. These and other principles are very applicable to LLMs and other advanced solvers, and shape our way of thinking what an algorithmic solver can or cannot do. In future work, we hope to apply these principles to such advanced algorithmic solvers, whose concepts of planning time and memory space may not be as simple to reason about.

\section*{Impact Statement}


This paper presents research aimed at advancing the field of Machine Learning. Our work may carry various potential societal implications, none of which we deem necessary to explicitly emphasize in this context.

\bibliography{bibliography}
\bibliographystyle{icml2024}

\newpage
\appendix
\onecolumn
\newpage
\section{Appendix - Training Setup}
\label{sec-train-setup}

In this work, we use various tasks to demonstrate our claims:
\begin{enumerate}
    \item \textbf{Graph shortest Path Task}
    This task has been used previously to evaluate DNC. In this task, the model sequentially receives a description of the graph $G(V,E)$ as its set of edges over $|E|$ steps, a query $(s,t)$ where $s$ is the source node and $t$ is the target node, and its output is a set of ordered edges describing the shortest path from $s$ to $t$.

    \item \textbf{MinCut Task}
    This task has not been used before as a an algorithmic task to the best of our knowledge, the model also receives a description of a connected graph $G(V,E)$ as its set of edges over $|E|$ steps, but no query.
    The output is a set of edges that describe the global minimum cut of the graph. As a reminder, a cut is a set of edges that once removed from the graph it will no longer be connected. A minimum cut is a cut with a minimum number of edges.

    \item \textbf{Associative Recall}
    A simple task previously used to test memory-augmented neural networks \cite{graves2014neural}, the model receives a list of $n$ items, where each item is a sequence of binary vectors. After the items are presented to the network, it receives a query of a random item and its output should be the subsequent item in the list.
    
    \item \textbf{Convex Hull}
    Given a set of $n$ 2D points coordinates, the model's goal is to find a set of points representing the smallest convex polygon containing all the points. 
\end{enumerate}

\subsection{Data Generation}
\label{sec-data-gen}
For the training process, we adopted a curriculum-based approach, training the model on increasingly larger samples and more complex queries. This meant parameterizing the dataset by input size.
Every 1000 training steps, the model is evaluated on the current lesson. If its accuracy exceeds 80\%, we move the training to the next lesson. The models trained on the final lesson for a constant pre-set number of steps, 

\subsubsection*{Shortest Path Task Graph Generation}
Each curriculum lesson is parameterized by the number of nodes $[n_1, n_2]$, average degree $[d_1, d_2]$, and path length $[p_1, p_2]$. The training graph are sampled uniformly from the set of all graphs with $n$ nodes and $m$ edges where $n$ is uniformly sampled from $[n_1, n_2]$,  and the number of edges $m$ is uniformly sampled from $\left[\lfloor \frac{N\cdot d_1}{2} \rfloor, \lfloor \frac{N\cdot d_2}{2} \rfloor\right]$.

\subsubsection*{MinCut Task Graph Generation}
Each curriculum lesson in this task is parameterized by the number of nodes $[n_1, n_2]$, clusters $[C_1, C_2]$, and cut size $[c_1, c_2]$.
The generator samples graphs as follows:
\begin{enumerate}
    \item 
    Split the $n$ nodes into $k$ disjoint groups, denoted as $C_1, ..., C_k$. Each group contains at least $c+2$ nodes to ensure a non-trivial minimum cut.

    \item 
    For each group $C_i$ (where $1 \leq i \leq k$), randomly sample a graph from the space of all graphs with $|C_i|$ nodes and a minimum degree of at least $c+1$.
    
    \item 
    Randomly add $c$ edges to connect $C_1$ to the rest of the nodes.
    
    \item 
    If there are more than two groups ($k > 2$), add edges between different clusters to ensure that each cluster is connected to at least $c$ edges, or $c+1$ if a unique minimum cut is required. 

\end{enumerate}

For the purpose of this work, we used a constant $[C_1, C_2] = [2,3]$, and constrained the maximum number of edges in the graph overall by adding an additional parameret max degree $[d_1, d_2]$

\subsubsection*{Associative Recall Task Data Generation}
Each curriculum lesson is parameterized by the number of items in the list $[n_1, n_2]$, as well as the number of digits per item $[d_1, d_2]$. The items are sampled uniformly from the range possible items, and separated by a space delimiter.

\subsubsection*{Convex Hull Task Data Generation}
Each curriculum lesson is parameterized by the number of 2D points $n\in [n1, n2]$. to generate the samples in each lesson, we follow a similar data generation process to the one described in \cite{vinyals2017pointer}, the $n$ points are sampled uniformly from $[0,1]\times [0,1]$, and sorted by the x-coordinates. The output is the convex hull starting at the point with the smallest x, counterclockwise.

\subsubsection{Target Consistency}
For the Shortest Path Task, given a single graph and a query, multiple valid shortest paths could exist. 
This creates a possibility of the sample input sample having different targets throughout the training, creating ambiguity within the training samples.
To overcome this problem, we used graphs with a unique shortest path for training purposes.
Each sampled graph was modified to ensure a unique solution. This was done by computing all possible shortest paths using breadth-first search and removing edges that disconnect all but one of the shortest paths.
The modified graphs were used for training purposes only. For model evaluation, the generated graphs were used as is, and the model was correct on a query if its prediction matched any of the valid shortest paths.
Similarly, for the MinCut Task, we face a similar problem, and trained similarly by generating graphs with a unique minimum cut for training, and use general graphs for evaluation purposes.

\subsubsection{Graph Representation}
For a sampled graph $G(V,E)$, each node in the graph is assigned a unique label sampled from $[1, N_{max}]$, where $N_{max}$ is the maximum number of nodes that the model supports. 
Each node label was encoded as a one-hot vector of size $N_{max}$, making the size of each input vector in the sequence $2\cdot N_{max} + 2$, including the edge as well as the $<eoi>, <ans>$ tokens.
The final graph description consists of the set of edges $x\in \R ^{|E|\times 2\cdot (N_{max} + 2)}$

\subsection{Training}
\label{sec-training}

As we explained in Section \ref{sec-dnc}, the input sequence is divided into several distinct phases: input description, query, planning, and answer phases. When training on the graph tasks, for a sampled graph $G(V,E)$ with $|V|=n$ nodes and $|E|=m $ edges, a sampled query $q$, and a set of a predefined number of planning steps $p$, the answer phase starts at $t_a = (m+1+p)$.
To train the model, the model's output is considered in the cost function solely during the answer phase.
During this phase, the model initially receives an answer cue indicating the start of this phase, and its output is utilized as feedback for the next step until the termination token is received.\\
The model has output nodes $2\cdot N_{max}$, which correspond to 2 softmax distributions over the two labels describing a single edge. Consequently, the log probability of correctly predicting the edge tuple is the sum of the log probabilities of correctly classifying each of the nodes.\\
For clarity, in the next section we denote the policy that the model learns over the actions $a\in \mathcal{A}$ by $\pi(a|s)$, where $s\in \mathcal{S}$ is the current state of the model. Additionally, we will refer to the correct answer sequence by $y = [y_1;...;y_{T}]$, where $T-1$ is the length of the shortest path. Lastly, the output of the model at each time step is denoted by $o_t = [o_t^1; o_t^2]$.

The cross-entropy loss corresponding to a single time step in the answer phase is:
\begin{equation*}
    \ell(o_t, y_t) = - \sum_{i=1}^2{\log{[\Pr (y_t|o^i_t)]}}
\end{equation*}
And the overall loss over the whole input sequence:
\begin{equation*}
    \mathcal{L} (o, y) = \sum_{t=0}^T{ \ell(o_{t+t_a}, y_t)}
\end{equation*}

In addition, we used teacher forcing to demonstrate optimal behaviour. This is commonly used in training recurrent neural networks that use the output of previous time steps as input to the model.
Since during the answer phase the model's prediction at time $t$ is a function of $o_{t-1}$, the teacher forcing provides the model with the correct answer instead of its own prediction allowing it to learn the next prediction based on the correct history. This helps in the early time steps when the model has not yet converged.
Formally, the current state during the answer phase $t > t_a$ is a function of the output of the model in the previous step $s_t = f(o_{t-1})$, and the next output is calculated as $o_t = \pi (\cdot |s_t)$. When the model is trained using teacher forcing only, the current loss is calculated as a function of the correct prediction in the previous step. Overall the loss is calculated as follows:
\begin{equation*}
    \mathcal{L} (\hat{o}, y) = \sum_{t=0}^T{ \ell(\hat{o}_{t+t_a}, y_t)]}
\end{equation*}

where:
\begin{equation*}
    \hat{o_t} = 
    \begin{cases}
    o_t & \text{if }t\leq t_a\\
    \pi(\cdot |f(y_{t-1})) & \text{otherwise}
    \end{cases}
\end{equation*}

In practice, we followed \cite{Graves2016}, and used a mixed training policy to guide the answer phase, by sampling from the optimal policy with probability $\beta$ and from the network prediction with probability $1-\beta$.

Finally, we trained using a memory $M\in \R ^{200\times 128}$ for the tasks of Shortest Path and Mincut, $M\in \R ^{100\times 32}$ for the Associative Recall task, and $M \in \R ^ {50\times 64}$ for the Convex Hull task.

\newpage
\section{Appendix - Curriculums}
\label{sec-curriculums}

\begin{table}[!htbp]
    \centering
    \begin{tabular}{|c|c|c|c|c|c|c|}
    \hline
     Lesson & Nodes & Average Degree & Path Length  \\
        \hline
        1 & (5,10)   & (1,2) & 2    \\
        2 & (5,20)   & (1,2) & 2   \\
        3 & (10,20)  & (1,2) & 2   \\
        4 & (10,20)  & (1,2) & (2,3)  \\
        5 & (10,20)  & (1,2) & (2,3)   \\
        6 & (10,20)  & (1,3) & (2,3)   \\
        7 & (10,20)  & (2,3) & (2,4)   \\
        8 & (10,20)  & (2,3) & (2,4)   \\
        9 & (10,25)  & (2,4) & (2,4)   \\
        10 & (10,25) & (2,4) & (2,5)  \\
        11 & (15,25) & (2,4) & (2,5)  \\
        12 & (15,25) & (2,5) & (2,5)  \\
        13 & (20,25) & (2,5) & (2,5)  \\
        14 & (20,25) & (2,6) & (2,5)  \\
        \hline
    \end{tabular}
    \caption{\textbf{Shortest Path Task curriculum} - parenthesis represent ranges (minimum value, maximum value).}
    \label{tab:supervised-10-linear}
\end{table}
\begin{table}[!htbp]
    \centering
    \begin{tabular}{|c|c|c|c|c|c|}
    \hline
     Lesson & Nodes & Cut Size & Max Degree  \\
            &       &          & Per Cluster\\
        \hline
        1 & (10,15)   & (1,1) & 3  \\
        2 & (10,15)   & (2,3) & 5  \\
        3 & (15,20)   & (2,3) & 5  \\
        4 & (15,20)   & (2,4) & 6   \\
        5 & (20,25)   & (2,4) & 6   \\
        \hline
    \end{tabular}
    \caption{\textbf{MinCut Task curriculum}}
    \label{tab:mincut-results}
\end{table}
\begin{table}[!htbp]
    \parbox{.45\linewidth}{
    \centering
    \begin{tabular}{|c|c|}
    \hline
     Lesson & Number   \\
            & of Items \\
        \hline
        1 & (5,10)   \\
        2 & (5,15)   \\
        3 & (10,20)   \\
        4 & (15,25)   \\
        5 & (20,25)   \\
        \hline
    \end{tabular}
    \caption{\textbf{Associative Recall Task curriculum} - The parameter $d\in [1,5]$ in all lessons.}
    }
    \hfill
    \parbox{.45\linewidth}{
    \centering
    \begin{tabular}{|c|c|}
    \hline
     Lesson & Number   \\
            & of Points \\
        \hline
        1 & (5,20)   \\
        2 & (10,20)  \\
        3 & (15,25)  \\
        4 & (20,35)  \\
        \hline
    \end{tabular}
    \caption{\textbf{Convex Hull Task curriculum}}
    }
\end{table}
\newpage
\section{Appendix - Stability and Stochastic Planning: Shortest Path}
\label{sec-bad-seeds}

Throughout this work, we encountered challenges related to the instability of DNC training, leading to instances of failure when evaluating most of the planning budgets across multiple seeds. In Section \ref{sec-stability}, we address this issue by proposing a solution that involves adding a stochastic number of planning steps. The effectiveness of this method is demonstrated on a model trained with an adaptive budget for the Associative Recall task. Here, we present the results of applying the stochastic fine-tuning approach to the Shortest Path task.

\begin{figure}[!htbp]
    \centering
    \begin{subfigure}{0.31\linewidth}
        \centering
        \includegraphics[width=\linewidth]{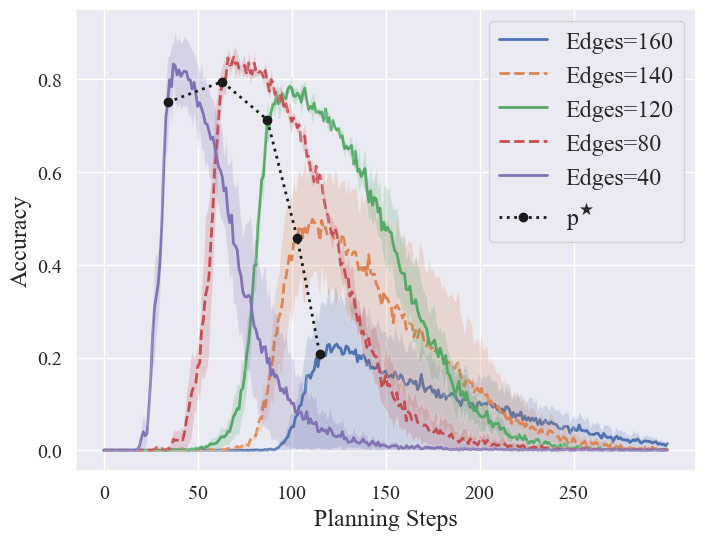}
        \caption{$A_n(p)$ for different input sizes $n$ before finetuning}
        \label{fig:sp-bad-peaks} 
    \end{subfigure}
    \hfill
    \begin{subfigure}{0.33\linewidth}
        \centering
    \includegraphics[width=\linewidth]{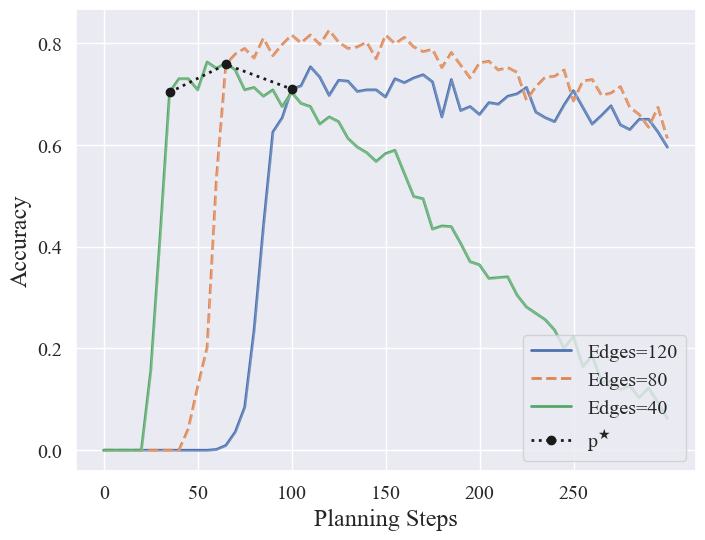}
        \caption{$A_n(p)$ for different input sizes $n$ - after stochastic finetuning}
        \label{fig:sp-bad-optimal}
    \end{subfigure}
    \begin{subfigure}{0.33\linewidth}
        \centering
    \includegraphics[width=\linewidth]{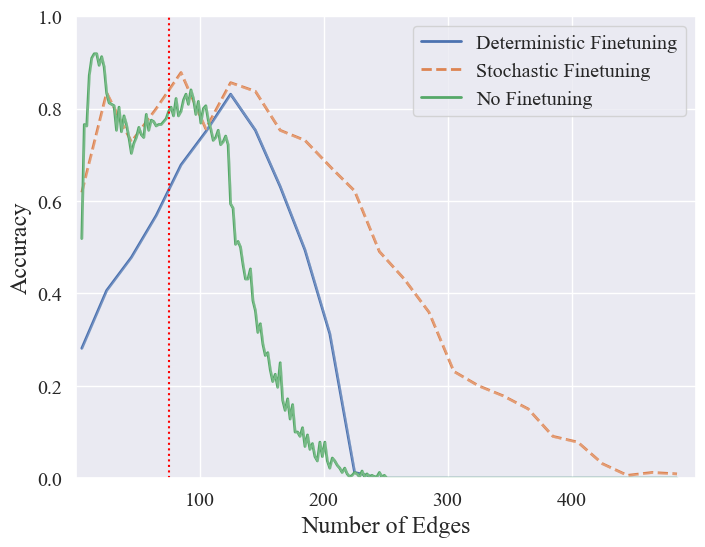}
        \caption{Effect of Stochastic Fine-tuning on Generalization}
        \label{fig:sp-bad-ext}
    \end{subfigure}
    \caption{
        \textbf{Empirical Results for ``failure cases'' when training with linear planning budget, Shortest Path.} While these models where trained with a linear planning budget, they did not learn to generalize well, and lack the ability to hold a steady state. The stochastic planning encourages the facilitation of a steady state, leading to better generalization}
    \label{fig:bad-seeds-peaks-optimal}
\end{figure}

In Section \ref{sec-generalization}, we showcased that training DNC models with a linear planning budget on the Shortest Path task enabled them to learn algorithms that generalize successfully to larger input sizes, outperforming constant planning budgets. However, some models trained under the exact same conditions turned out to be failure cases, exhibiting behavior similar to the baseline models with a constant budget. As illustrated in Figure \ref{fig:sp-bad-peaks}, these models are sensitive to changes in the planning parameter $p$ and fail to maintain a steady state when tested on larger inputs.

We applied the proposed approach of introducing a stochastic number of planning steps to address these issues. Figure \ref{fig:sp-bad-ext} clearly shows that after fine-tuning the model with the addition of stochastic planning steps, the model demonstrates improved generalization to longer inputs. Deterministic fine-tuning, on the other hand, made little difference. Additionally, Figure \ref{fig:sp-bad-optimal} reveals that the models are now much less sensitive to variations in the planning phase duration, exhibiting steady-state behavior.

\newpage
\section{Appendix - Quadratic Planning Budget Results}
\label{sec-quadratic}

We Supply initial results with a quadratic planning budget $p=n^2$ trained on Graph Shortest Path. This model require a very long training time and as a result, we had to cut short the experiment after a mere 3\% of the training process on the final lesson. We stopped the model after 3K steps in the last lesson, while for the other budgets we performed 100K.

\begin{figure}[!htbp]
    \centering
    \begin{subfigure}{0.49\linewidth}
        \centering
        \includegraphics[width=\linewidth]{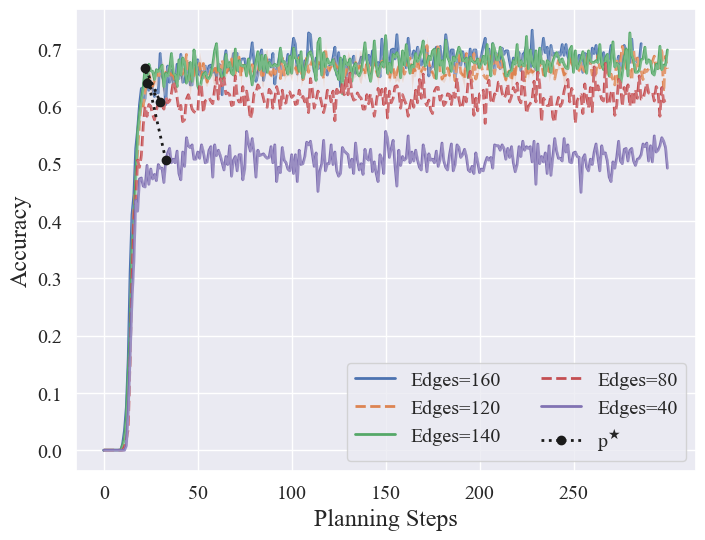}
        \caption{Model accuracy when trained with $p(n)=\abs{E}^2$}
        \label{fig:sp-quad-peaks} 
    \end{subfigure}
    \hfill
    \begin{subfigure}{0.49\linewidth}
        \centering
    \includegraphics[width=\linewidth]{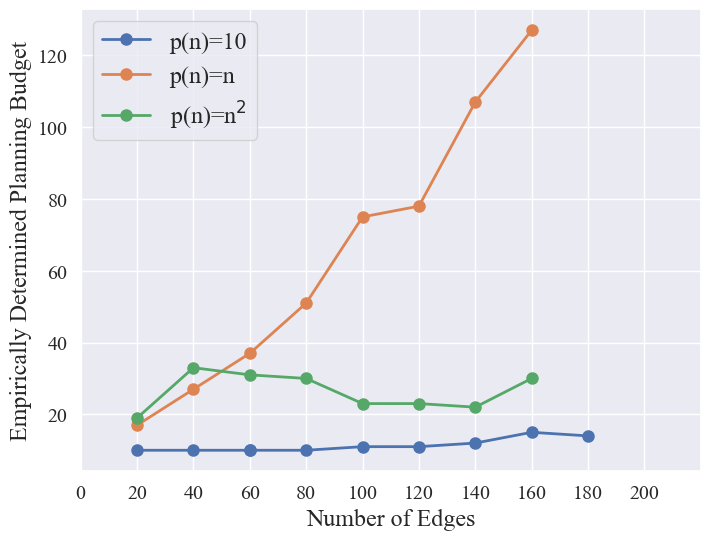}
        \caption{Empirically Determined Planning Budget}
        \label{fig:sp-quad-optimal}
    \end{subfigure}
    \\[\baselineskip]
    \begin{subfigure}{0.49\linewidth}
        \centering
    \includegraphics[width=\linewidth]{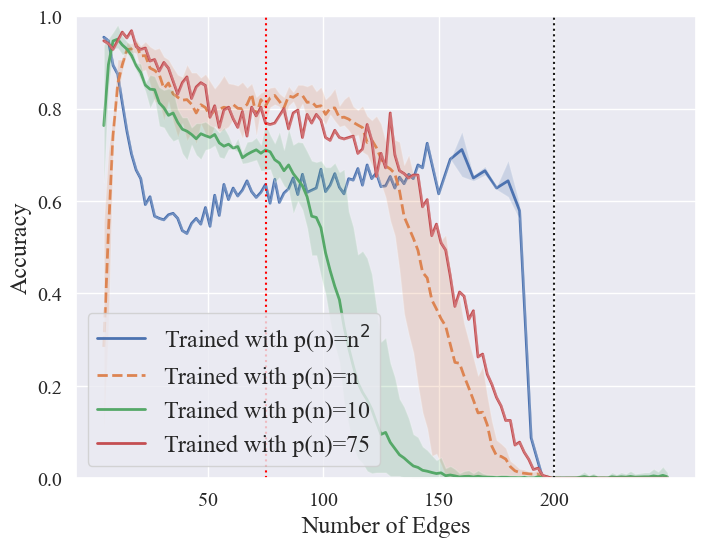}
        \caption{Generalization to input sizes not seen during training}
        \label{fig:quad-generalize}
    \end{subfigure}
    \hfill
    \begin{subfigure}{0.49\linewidth}
        \centering
    \includegraphics[width=\linewidth]{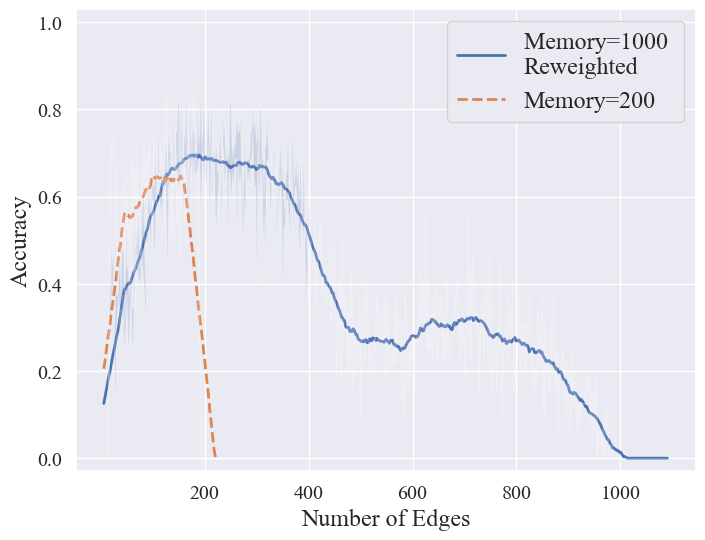}
        \caption{Generalization with memory extension}
        \label{fig:sp-quad-infer-mem-extension}
    \end{subfigure}
    \\[\baselineskip]
    \caption{
        \textbf{Empirical Results for the partially trained model that trained with a quadratic planning budget.} It is evident that even with 3\% of training time, the model surpasses smaller budgets when generalizing to larger input sizes.}
    \label{fig:quad-bad-seeds-peaks-optimal}
\end{figure}

Even with this extreme setback, a model that applies a quadratic planning budget reaches higher accuracy for large inputs, compared to models that use smaller budgets, as can be seen in Figure \ref{fig:quad-generalize}. Furthermore, it appears that although the model uses a quadratic number of planning steps, it reaches the final steady state described in Section \ref{sec-empirical-budget} very quickly. A nearly constant number of planning steps is needed, as can be seen in figure \ref{fig:sp-quad-peaks},\ref{fig:sp-quad-optimal}. As Shortest Path can be solved in linear time, we hypothesize this model was able to learn a very efficient representation space while still learning to use it's memory.
This could be explained by the very large duration of the planning phase: for most input sizes during training, the model really doesn't need most of the planning phase it is give. This forces it to become very good at holding the steady state it reaches after finding the final answer, which perhaps allows it to then further optimize its representation. This in turn reduces its actual runtime considerably.

Furthermore, we show preliminary results for the generalization of this model in Figure \ref{fig:sp-quad-infer-mem-extension}.
This shows even better generalization performance than we saw with the linear budget model, which is very promising.

Unfortunately the very long runtimes become prohibitive to train such a model. This points at a limit of solving such problems using DNCs - complex problems will require a large amount of planning steps to train.

\newpage
\section{Appendix - MinCut Supplementary Figures}
\label{sec-mincut}

\begin{figure}[!htbp]
    \centering
    \begin{subfigure}{0.32\linewidth}
        \centering
        \includegraphics[width=\linewidth]{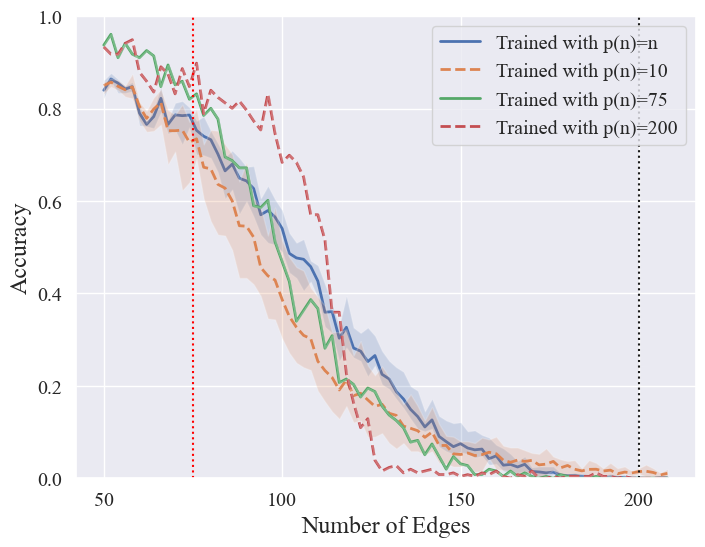}
        \caption{Generalization}
        \label{fig:mc-gen}
    \end{subfigure}
    \hfill
    \begin{subfigure}{0.32\linewidth}
        \centering
        \includegraphics[width=\linewidth]{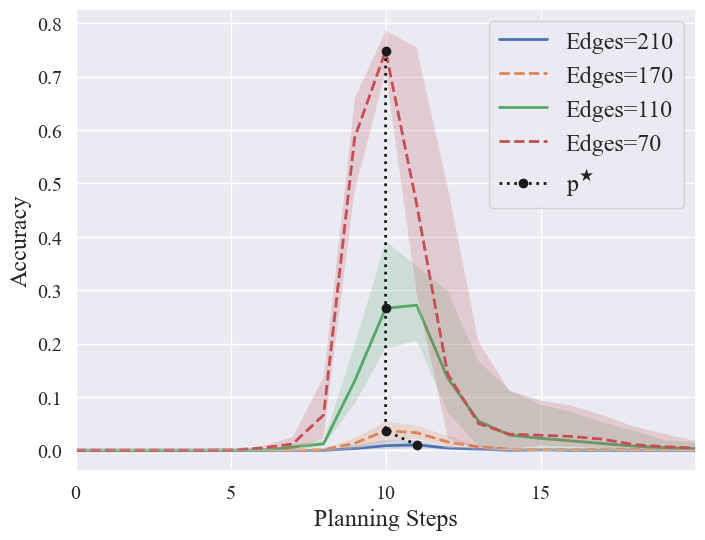}
        \caption{Model accuracy when trained with $p(n)=10$}
        \label{fig:mc-constant-peaks} 
    \end{subfigure}
    \hfill
    \begin{subfigure}{0.32\linewidth}
        \centering
        \includegraphics[width=\linewidth]{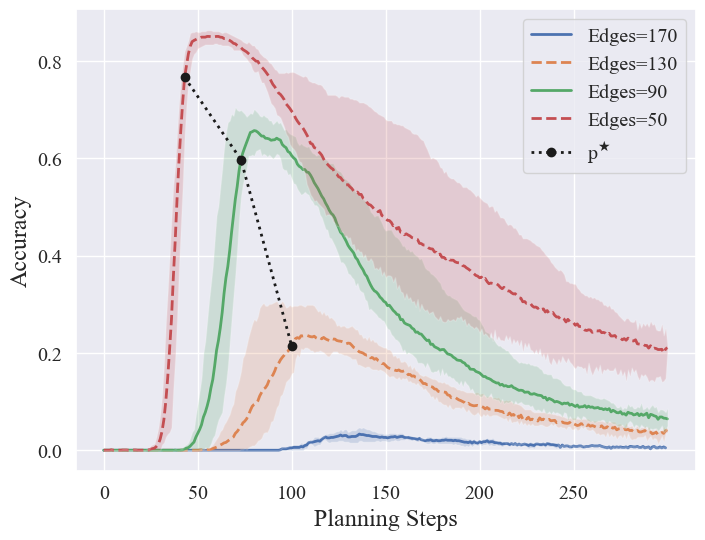}
        \caption{Model accuracy when trained with $p(n)=n$}
        \label{fig:mc-linear-peaks}
    \end{subfigure}
    \\[\baselineskip]
    \begin{subfigure}{0.32\linewidth}
        \centering
        \includegraphics[width=\linewidth]{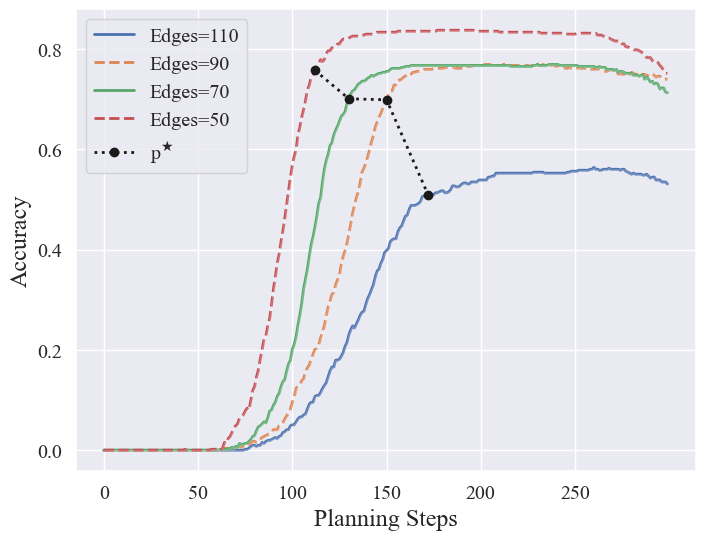}
        \caption{Model accuracy when trained with $p(n)=200$}
        \label{fig:mc-constant200-peaks} 
    \end{subfigure}
    \hfill
    \begin{subfigure}{0.32\linewidth}
        \centering
        \includegraphics[width=\linewidth]{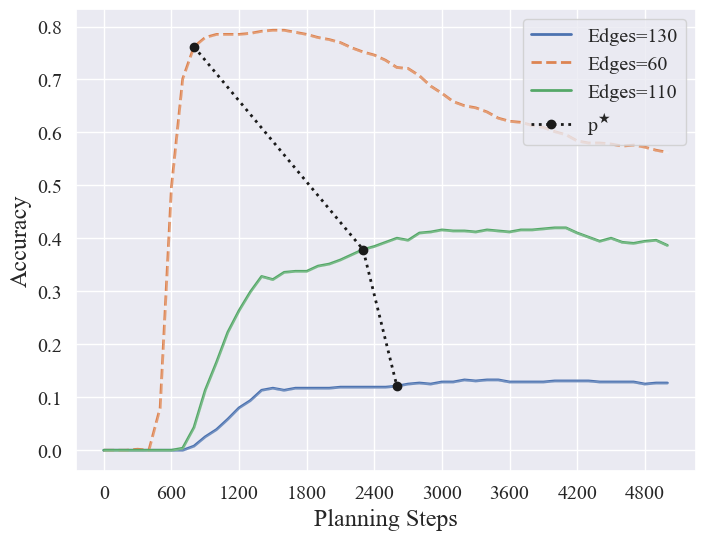}
        \caption{Model accuracy when trained with $p(n)=|V||E|$}
        \label{fig:mc-VE-peaks}
    \end{subfigure}
    \hfill
    \begin{subfigure}{0.32\linewidth}
        \centering
        \includegraphics[width=\linewidth]{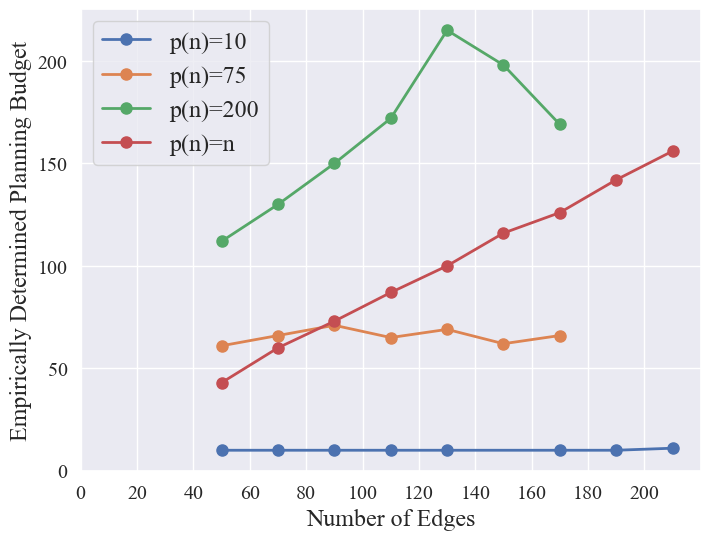}
        \caption{Empirically Determined Planning Budget}
        \label{fig:mc-opt}
    \end{subfigure}
    \\[\baselineskip]
    \caption{
        \textbf{Empirical Results, MinCut} - Planning budget drastically changes model behavior. Although good generalization was not achieved, which we attribute to problem complexity being too hard for the DNC, we do notice interesting behaviors, including a steady-state with a large constant budget, and non trivial learned time complexity.
    }
    \label{fig:mincut-peaks} 
\end{figure}

We supply additional results for solving the MinCut task with different planning budgets. As can be seen in Figure \ref{fig:mincut-peaks}, we again prove allowing a larger constant budget or an adaptive one result in very different behaviors compared to the baselines, who fails to generalize to larger inputs. 

Unfortunately, training a model with a planning budget of size $\abs{V}\abs{E}$ proved to be a very long, hard and unstable process, which is expected as MinCut'S time complexity requires the recurrent neural network to process extremely long sequences.

\newpage
\section{Appendix - Associative Recall Supplementary Material}
\label{sec-ass}
\begin{figure}[!htbp]
    \centering
    \begin{subfigure}{0.33\linewidth}
        \centering
        \includegraphics[width=\linewidth]{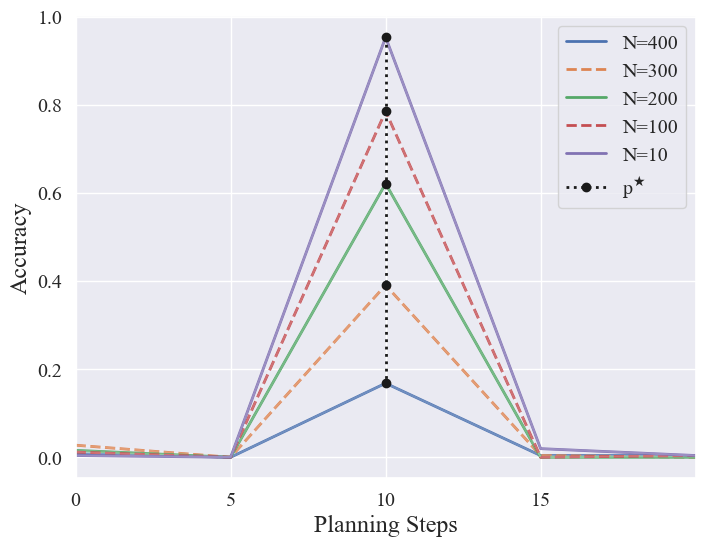}
        \caption{Model accuracy when trained with $p(n)=10$}
        \label{fig:ar-10-peaks} 
    \end{subfigure}
    \hfill
    \begin{subfigure}{0.33\linewidth}
        \centering
        \includegraphics[width=\linewidth]{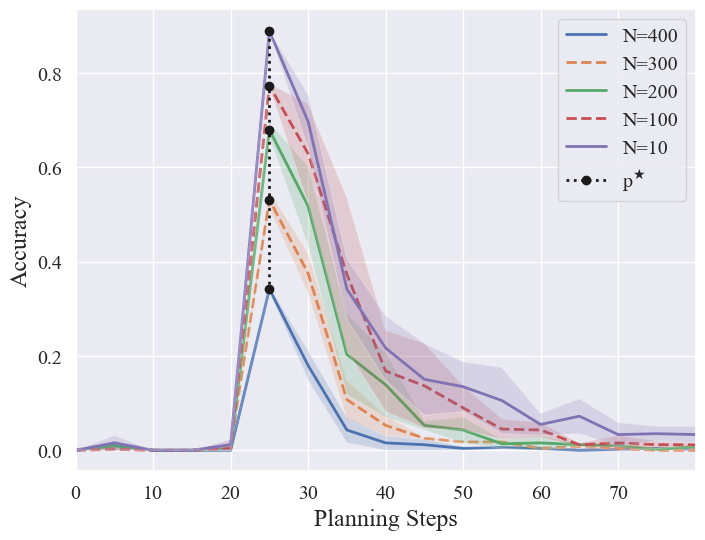}
        \caption{Model accuracy when trained with $p(n)=25$}
        \label{fig:ar-25-peaks}
    \end{subfigure}
    \hfill
    \begin{subfigure}{0.33\linewidth}
        \centering
        \includegraphics[width=\linewidth]{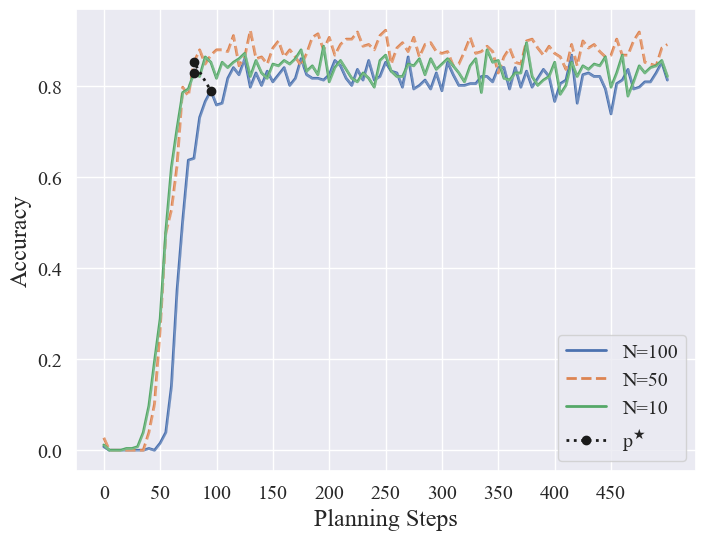}
        \caption{Model accuracy when trained with $p(n)=100$}
        \label{fig:ar-100-peaks} 
    \end{subfigure}
    \\[\baselineskip]
    \begin{subfigure}{0.33\linewidth}
        \centering
        \includegraphics[width=\linewidth]{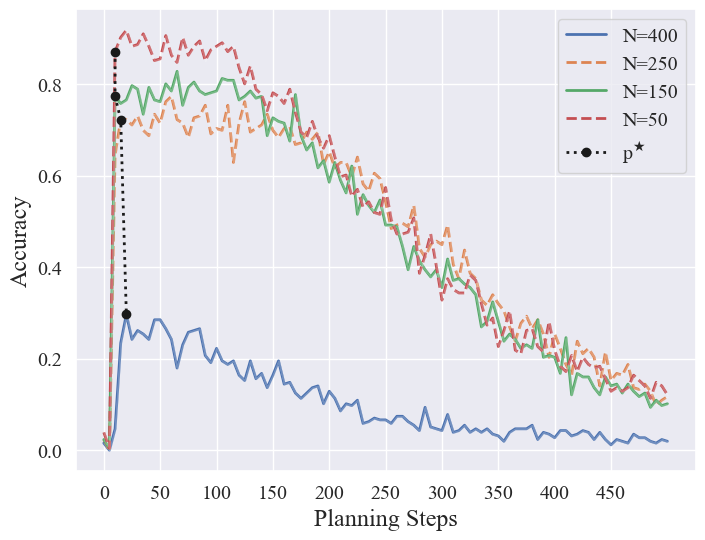}
        \caption{Model accuracy when trained with $p(n)=n$}
        \label{fig:ar-linear-peaks}
    \end{subfigure}
    \quad
    \begin{subfigure}{0.33\linewidth}
        \centering
    \includegraphics[width=\linewidth]{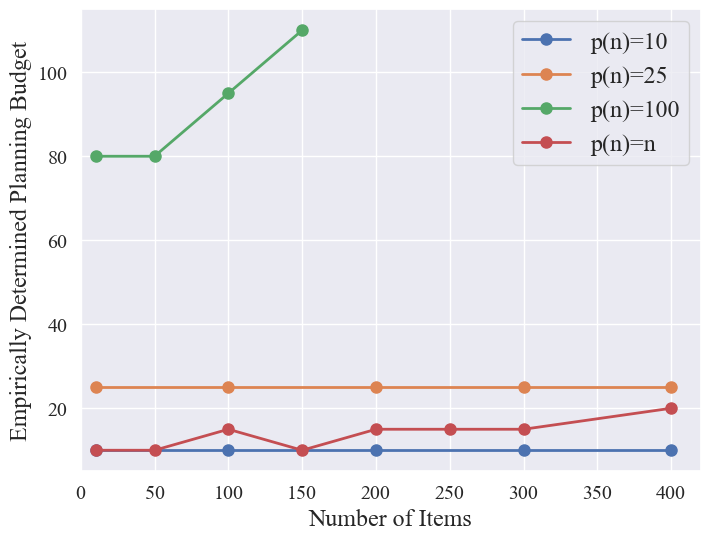}
        \caption{Empirically Determined Planning Budget for Associative Recall}
        \label{fig:ar-optimal}
    \end{subfigure}
    \\[\baselineskip]
    \caption{
        \textbf{$A_n(p)$ for different input sizes $n$, Associative Recall} - The baseline found an online solution and additional planning proved unnecessary or even harmful, categorizing the problem as easy. Interestingly, \textbf{(e)} shows that the adaptive model has learned an online algorithm with constant latency as well, although less effective than the baseline.
    }
    \label{fig:ar-peaks} 
\end{figure}

We supply additional results for solving the Associative Recall Task with different planning budgets. As can be seen in Figure \ref{fig:ar-peaks}, all budgets appear to find a solution in relatively short time. The small constants are unable to utilize additional planning time, as they found a solution online. The linear budget also found a solution quickly. The large constant found a solution and when given more planning time simply holds it in memory, as the graph plateaus.  The latter two are negligible however, as we have seen in Section \ref{sec-exp} that they fail to generalize to larger inputs, which we attribute to the multiple problems caused by a planning budget that is simply too large, such as instability and vanishing gradients. 


We also note that the memory complexity of the problem is linear as the data has to be saved to memory, hence using an adaptive memory $m(n)$ was crucial for this problem just like the others.  

\newpage
\section{Appendix - Convex Hull Supplementary Material}
\label{sec-convex}
\begin{figure}[!htbp]
    \centering
    \begin{subfigure}{0.33\linewidth}
        \centering
        \includegraphics[width=\linewidth]{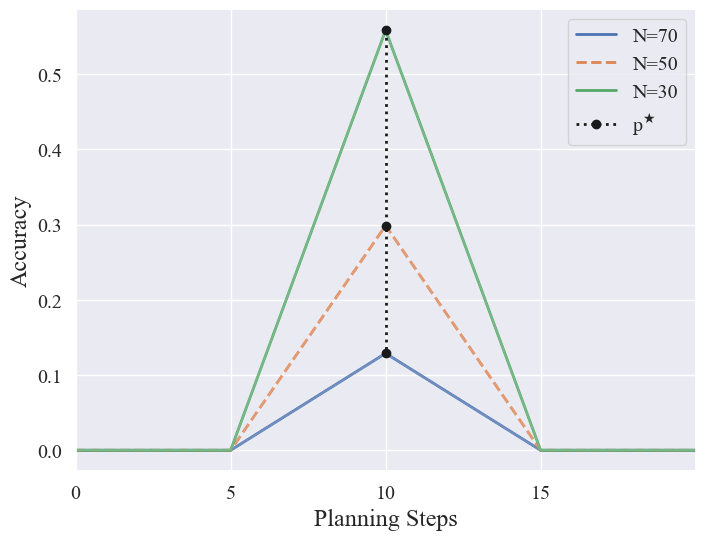}
        \caption{Model accuracy when trained with $p(n)=10$}
        \label{fig:ch-10-peaks} 
    \end{subfigure}
    \hfill
    \begin{subfigure}{0.33\linewidth}
        \centering
        \includegraphics[width=\linewidth]{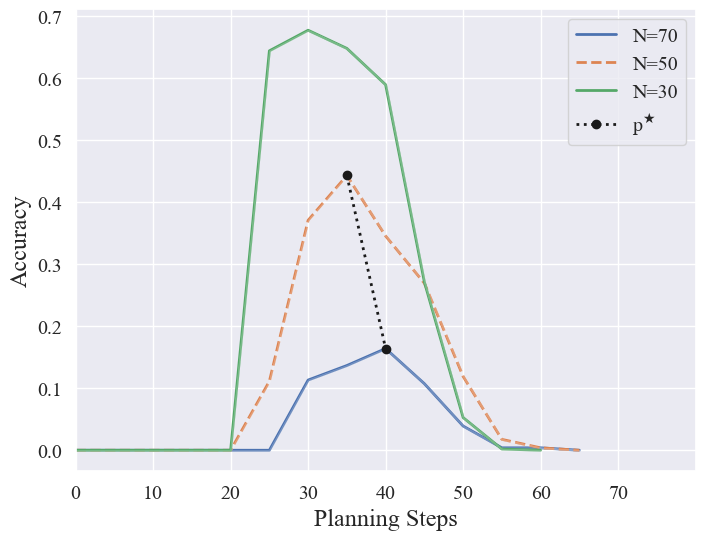}
        \caption{Model accuracy when trained with $p(n)=25$}
        \label{fig:ch-25-peaks}
    \end{subfigure}
    \hfill
    \begin{subfigure}{0.33\linewidth}
        \centering
        \includegraphics[width=\linewidth]{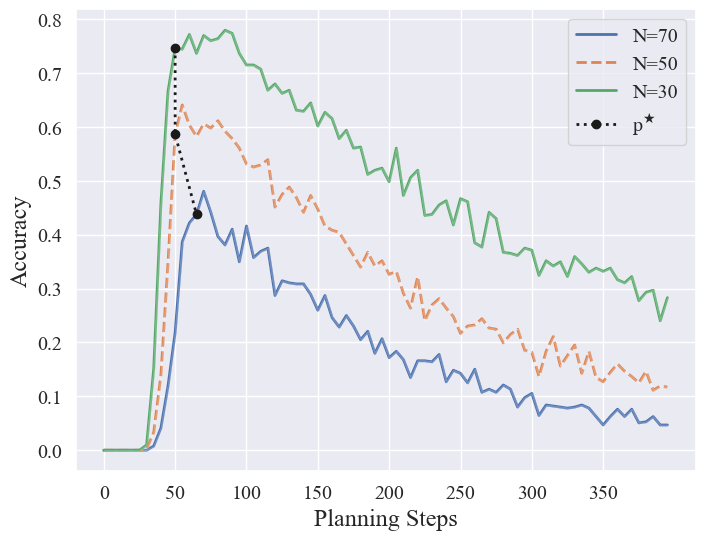}
        \caption{Model accuracy when trained with $p(n)=50$}
        \label{fig:ch-50-peaks} 
    \end{subfigure}
    \\[\baselineskip]
    \begin{subfigure}{0.33\linewidth}
        \centering
        \includegraphics[width=\linewidth]{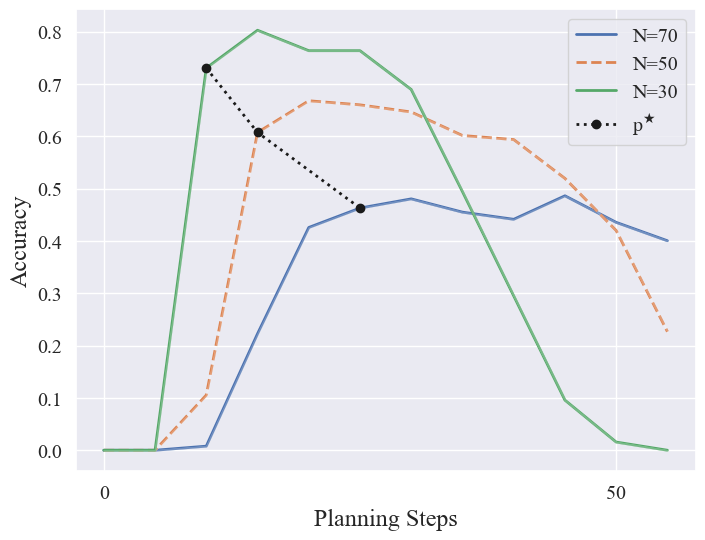}
        \caption{Model accuracy when trained with $p(n)=n/2$}
        \label{fig:ch-linear-peaks-n2}
    \end{subfigure}
    \hfill
    \begin{subfigure}{0.33\linewidth}
        \centering
        \includegraphics[width=\linewidth]{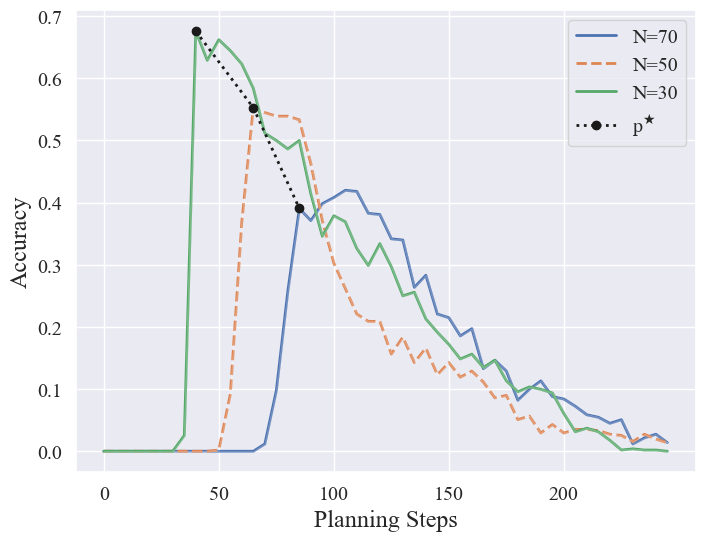}
        \caption{Model accuracy when trained with $p(n)=3n/2$}
        \label{fig:ch-linear-peaks-3n2}
    \end{subfigure}
    \hfill
    \begin{subfigure}{0.33\linewidth}
        \centering
    \includegraphics[width=\linewidth]{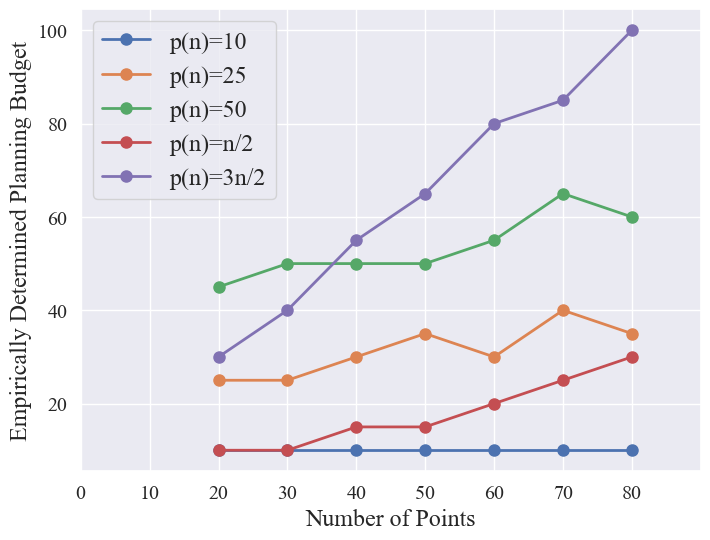}
        \caption{Empirically Determined Planning Budget}
        \label{fig:ch-optimal}
    \end{subfigure}
    \\[\baselineskip]
    \begin{subfigure}{0.33\linewidth}
        \centering
        \includegraphics[width=\linewidth]{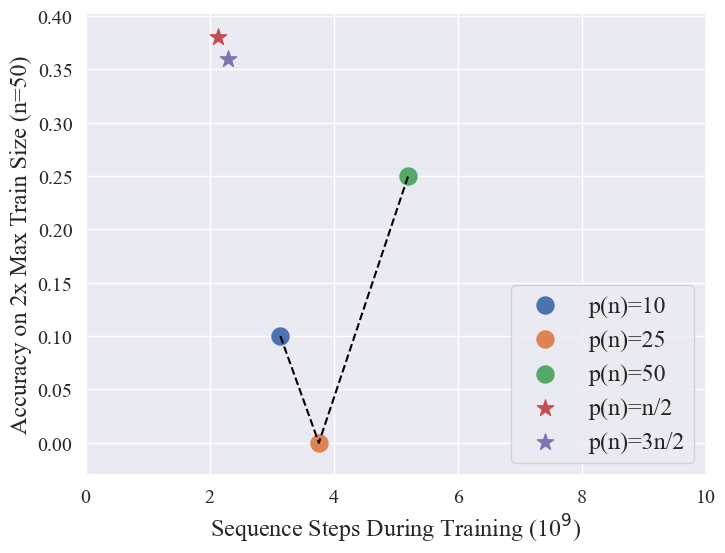}
        \caption{Training FLOPs}
        \label{fig:ch-flops}
    \end{subfigure}
    \\[\baselineskip]
    \caption{
        \textbf{$A_n(p)$ for different input sizes $n$, Convex Hull} - The adaptive model outperforms the constant baseline, and is nearly matched only by the largest constant budget of $p(n)=50$. \textbf{(f)} shows that the complexity of the models matches the training planning budget. Additionally, Figure \textbf{(g)} demonstrates the efficiency of the adaptive planning budget, as it requires the least training FLOPs, while generalizing best to larger inputs.
    }
    \label{fig:ch-peaks} 
\end{figure}

We supply additional results for solving the Convex Hull task with different planning budgets. As can be seen in Figures \ref{fig:ch-linear-peaks-n2} and \ref{fig:ch-linear-peaks-3n2}, these models did not reach a steady state as the one reached in the Shortest Path task. We believe these models could reach a steady state and generalize better if allowed more training time. 

\end{document}